\documentclass[]{fairmeta} %

\usepackage{amsmath,amsfonts,bm}

\def\eqref#1{equation~\ref{#1}}

\def\1{\bm{1}}

\DeclareMathAlphabet{\mathsfit}{\encodingdefault}{\sfdefault}{m}{sl}
\SetMathAlphabet{\mathsfit}{bold}{\encodingdefault}{\sfdefault}{bx}{n}

\usepackage[utf8]{inputenc}  %
\usepackage[T1]{fontenc}     %
\usepackage{hyperref}        %
\usepackage{url}             %
\usepackage{booktabs}        %
\usepackage{amsfonts}        %
\usepackage{nicefrac}        %
\usepackage{microtype}       %
\usepackage{graphicx}
\usepackage{multicol}
\usepackage{multirow}
\usepackage{xspace}
\usepackage{wrapfig}
\usepackage{enumitem}
\usepackage[dvipsnames,table]{xcolor}
\usepackage{colortbl}
\usepackage{caption}
\usepackage{lmodern} 
\usepackage{algorithm}
\usepackage{algorithmic}
\usepackage{subcaption}
\usepackage{wrapfig}
\usepackage{xspace}
\usepackage{needspace}
\usepackage{varwidth}

\usepackage{multicol}
\usepackage{multirow}
\usepackage{pifont}
\usepackage{tabularx}
\newcolumntype{R}{>{\raggedleft\arraybackslash}X}
\definecolor{lightblue}{RGB}{220,235,250}
\definecolor{lightgray}{gray}{0.91}

\usepackage[most,skins,theorems]{tcolorbox}
\usepackage{listings}
\newtcolorbox{promptbox}[1][]{
  enhanced, breakable,
  colback=gray!1,      
  colframe=gray!60,    
  coltitle=black,      
  boxrule=2pt,
  arc=10pt,
  left=6pt, right=6pt, top=6pt, bottom=6pt,
  title={#1}, fonttitle=\bfseries,
  attach boxed title to top left={yshift*=-3mm},
  boxed title style={colback=gray!10}
}

\tcbset{
  aibox/.style={
    width=\linewidth,
    top=8pt,
    bottom=4pt,
    colback=inftythink-red!15,
    colframe=inftythink-red,
    colbacktitle=inftythink-red!90!black,
    enhanced,
    center,
    attach boxed title to top left={yshift=-0.1in,xshift=0.15in},
    boxed title style={boxrule=0pt,colframe=white,},
  }
}
\newtcolorbox{AIbox}[2][]{aibox,title=#2,#1}

\usepackage{amsthm}
\theoremstyle{plain}

\theoremstyle{definition}

\theoremstyle{remark}

\newcommand{\methodname}{Code-A1\xspace}

\title{
\methodname: Adversarial Evolving of Code LLM and Test LLM via Reinforcement Learning
}

\author[1 *]{Aozhe Wang}
\author[1 *]{Yuchen Yan}
\author[1 *]{Nan Zhou}
\author[1 *]{Zhengxi Lu}
\author[1]{Weiming Lu}
\author[1]{Jun Xiao}
\author[1]{Yueting Zhuang}
\author[1 \dagger]{Yongliang Shen}

\affiliation[1]{Zhejiang University}

\contribution[*]{Equal contributions}
\contribution[\dagger]{Corresponding author}

\abstract{
  Reinforcement learning for code generation relies on verifiable rewards from unit test pass rates. Yet high-quality test suites are scarce, existing datasets offer limited coverage, and static rewards fail to adapt as models improve. Recent self-play methods unify code and test generation in a single model, but face a inherent dilemma: white-box access leads to self-collusion where the model produces trivial tests for easy rewards, yet black-box restriction yields generic tests that miss implementation-specific bugs. We introduce \textbf{\methodname}, an adversarial co-evolution framework that jointly optimizes a Code LLM and a Test LLM with opposing objectives. The Code LLM is rewarded for passing more tests, while the Test LLM is rewarded for exposing more defects. This architectural separation eliminates self-collusion risks and safely enables white-box test generation, where the Test LLM can inspect candidate code to craft targeted adversarial tests. We further introduce a Mistake Book mechanism for experience replay and a composite reward balancing test validity with adversarial difficulty. Experiments on Qwen2.5-Coder models demonstrate that \methodname achieves code generation performance matching or exceeding models trained on human-annotated tests, while significantly improving test generation capability. 
}

\date{\today}
\metadata[Project Page]{\url{https://zju-real.github.io/Code-A1}}
\metadata[Code]{\url{https://github.com/ZJU-REAL/Code-A1}}
\correspondence{\email{\{waz,yanyuchen,syl\}@zju.edu.cn}}

\begin{document}

\usetikzlibrary{patterns,patterns.meta,positioning,backgrounds,calc}

\definecolor{mila-purple}{RGB}{102,46,125}
\definecolor{mila-khaki}{RGB}{245,231,206}
\definecolor{mila-khaki-dark}{RGB}{185,174,154}
\definecolor{mila-yellow}{HTML}{F9B40D}
\definecolor{mila-yellow-dark}{HTML}{DD950B}
\definecolor{mila-greenblue}{RGB}{130,207,190}
\definecolor{plot-blue}{RGB}{132,182,206}
\definecolor{plot-red}{RGB}{248,191,199}
\definecolor{plot-green}{RGB}{157,207,127}
\definecolor{mila-purple-1}{HTML}{EFE8F1}
\definecolor{mila-purple-2}{HTML}{D8C9DD}
\definecolor{mila-purple-3}{HTML}{C1AACA}
\definecolor{mila-purple-4}{HTML}{AA8BB7}
\definecolor{mila-purple-5}{HTML}{936BA3}
\definecolor{mila-purple-6}{HTML}{7C4D90}
\definecolor{mila-purple-dark}{HTML}{451059}
\definecolor{mila-purple-superdark}{HTML}{340040}
\definecolor{delethink-blue}{HTML}{3ba1ff}
\definecolor{delethink-purple}{HTML}{6441D2}
\definecolor{delethink-dark-purple}{HTML}{3C2880}

\definecolor{inftythink-red}{RGB}{248,191,199}
\definecolor{inftythink-blue}{RGB}{150,184,243}
\definecolor{inftythink-green}{RGB}{154,226,192}
\definecolor{inftythink-yellow}{RGB}{249,180,13}

\newlength{\sz}         \setlength{\sz}{0.7cm}
\newlength{\sw}         \setlength{\sw}{0.4cm}
\newlength{\sww}        \setlength{\sww}{0.37cm}
\newlength{\rectr}         \setlength{\rectr}{5pt}
\newlength{\rectw}         \setlength{\rectw}{2cm}
\newlength{\rectwconclu}   \setlength{\rectwconclu}{1cm}
\newlength{\rectwlongcot}  \setlength{\rectwlongcot}{15cm}
\newlength{\segap}      \setlength{\segap}{1.2cm}
\newlength{\segaparrow} \setlength{\segaparrow}{0.1cm}

\newlength{\BORDER}
\setlength{\BORDER}{1pt}

\tikzset{
  border/.style={line width=\BORDER},
  centerline/.style={dash pattern=on 1pt off 2pt, line cap=round, line width=\BORDER},
  arrowline/.style={->, line width=\BORDER},
  curvedarrow/.style={->, draw=black!30, line width=0.7pt, shorten >=2pt, shorten <=4pt}
}

\tikzset{
  segment marker shape/.is choice,
  segment marker shape/circle/.code={%
    \def\drawmarker##1{\fill[white] (##1) circle[radius=1.2pt];}%
  },
  segment marker shape/square/.code={%
    \def\drawmarker##1{\fill[white]
      ($(##1)+(-1.1pt,-1.1pt)$) rectangle ($(##1)+(1.1pt,1.1pt)$);}%
  },
  segment marker shape/diamond/.code={%
    \def\drawmarker##1{\fill[white]
      ($(##1)+(0,1.4pt)$)--($(##1)+(1.4pt,0)$)--($(##1)+(0,-1.4pt)$)--($(##1)+(-1.4pt,0)$)--cycle;}%
  },
  segment marker shape/triangle/.code={%
    \def\drawmarker##1{\fill[white]
      ($(##1)+(-1.2pt,-1.2pt)$)--($(##1)+(-1.2pt,1.2pt)$)--($(##1)+(1.6pt,0)$)--cycle;}%
  },
  segment marker shape/trangle/.style = {segment marker shape/triangle},
}

\tikzset{
  pics/inftythink_iter_1/.style n args={3}{
    code={
      \begin{scope}[node distance=0pt]
        \begin{scope}[local bounding box=sq]
          \path[
            draw=#1, line width=1.4\BORDER, dashed,
            preaction={fill=#1!5}, pattern color=black!30
          ]
          (\rectr,0) -- (\sw,0) -- (\sw,\sz) -- (\rectr,\sz)
          arc (90:180:\rectr) -- (0,\rectr) arc (180:270:\rectr) -- cycle;
        \end{scope}
        \node[font=\bfseries\large] at (sq.center) {$\mathrm{q}$};

        \begin{scope}[local bounding box=rect, xshift=3pt+\sw]
          \path[
            draw=#2, line width=1.4\BORDER,
            preaction={fill=#2!10}, pattern color=black!30
          ]
          (0.8\rectw,0) -- (0,0) -- (0,\sz) -- (0.8\rectw,\sz);
        \end{scope}

        \begin{scope}[xshift=3pt+\sw+0.8\rectw, local bounding box=summary]
          \path[
            draw=#3, line width=1.4\BORDER,
            preaction={fill=#3!10}, pattern color=black!30
          ]
          (0,\sz) -- (0.4\rectw,\sz) -- (0.4\rectw,0) -- (0,0);
        \end{scope}

        \begin{scope}[on background layer]
          \fill[#2!10] (rect.south west) rectangle (rect.north east);
          \fill[#3!10] (summary.south west) rectangle (summary.north east);
        \end{scope}

        \begin{scope}
          \clip (rect.south west) rectangle (rect.north east);
        
          \def\dia{1.8pt}
        
          \coordinate (gxW) at ($(rect.west)!0.19!(rect.east)$);
          \coordinate (gxE) at ($(rect.west)!0.81!(rect.east)$);
        
          \coordinate (gy1) at ($(rect.south)!0.75!(rect.north)$);
          \coordinate (p1)  at ($(gxW|-gy1)$);
          \fill[#2!60]
            ($(p1)+(-\dia,0)$) -- ($(p1)+(0,\dia)$) -- ($(p1)+(\dia,0)$) -- ($(p1)+(0,-\dia)$) -- cycle;
          \draw[#2!60, line width=\BORDER, line cap=round]
            ($(p1)+(\dia+0.6pt,0)$) -- ([xshift=-2pt]gxE|-gy1);
        
          \coordinate (gy2) at ($(rect.south)!0.50!(rect.north)$);
          \coordinate (p2)  at ($(gxW|-gy2)$);
          \fill[#2!60]
            ($(p2)+(-\dia,0)$) -- ($(p2)+(0,\dia)$) -- ($(p2)+(\dia,0)$) -- ($(p2)+(0,-\dia)$) -- cycle;
          \draw[#2!60, line width=\BORDER, line cap=round]
            ($(p2)+(\dia+0.6pt,0)$) -- ([xshift=-5pt]gxE|-gy2);
        
          \coordinate (gy3) at ($(rect.south)!0.25!(rect.north)$);
          \coordinate (p3)  at ($(gxW|-gy3)$);
          \fill[#2!60]
            ($(p3)+(-\dia,0)$) -- ($(p3)+(0,\dia)$) -- ($(p3)+(\dia,0)$) -- ($(p3)+(0,-\dia)$) -- cycle;
          \draw[#2!60, line width=\BORDER, line cap=round]
            ($(p3)+(\dia+0.6pt,0)$) -- (gxE|-gy3);
        \end{scope}

        \begin{scope}
          \clip (summary.south west) rectangle (summary.north east);
        
          \path coordinate (gxW) at ($(summary.west)!0.18!(summary.east)$);
          \path coordinate (gxE) at ($(summary.west)!0.75!(summary.east)$);
        
          \path coordinate (gy1) at ($(summary.south)!0.75!(summary.north)$);
          \fill[#3!60] (gxW|-gy1) circle[radius=1.2pt];
          \draw[#3!60, line width=\BORDER, line cap=round]
               ($(gxW|-gy1)+(1.6pt,0)$) -- ([xshift=-0pt]gxE|-gy1);
        
          \path coordinate (gy2) at ($(summary.south)!0.50!(summary.north)$);
          \fill[#3!60] (gxW|-gy2) circle[radius=1.2pt];
          \draw[#3!60, line width=\BORDER, line cap=round]
               ($(gxW|-gy2)+(1.6pt,0)$) -- ([xshift=-3pt]gxE|-gy2);
        
          \path coordinate (gy3) at ($(summary.south)!0.25!(summary.north)$);
          \fill[#3!60] (gxW|-gy3) circle[radius=1.2pt];
          \draw[#3!60, line width=\BORDER, line cap=round]
               ($(gxW|-gy3)+(1.6pt,0)$) -- ([xshift=-6pt]gxE|-gy3);
        \end{scope}

        \coordinate (-sqwest)   at (sq.west);
        \coordinate (-rectwest) at (rect.west);
        \coordinate (-recteast) at (rect.east);
        \coordinate (-east)     at (summary.east);
        \coordinate (-north) at ([xshift=1pt]summary.north);
        \coordinate (-rectnorth) at ([xshift=0.25\rectw]rect.north);
        \coordinate (-rectsouth) at (rect.south);
      \end{scope}
    }
  },
  pics/inftythink_iter_1/.default={mila-purple-1}{mila-purple-2}
}

\tikzset{
  pics/inftythink_iter_n-1/.style n args={3}{
    code={
      \begin{scope}[node distance=0pt]
        \begin{scope}[local bounding box=sq]
          \path[draw=none, preaction={fill=#1!10}, pattern color=black!30]
            (\rectr,0) -- (\sww,0) -- (\sww,\sz) -- (\rectr,\sz)
            arc (90:180:\rectr) -- (0,\rectr) arc (180:270:\rectr) -- cycle;
          \draw[#1, dashed, line width=1.4\BORDER]
            (\sww,\sz) -- (\rectr,\sz) arc (90:180:\rectr) -- (0,\rectr)
            arc (180:270:\rectr) -- (\rectr,0) -- (\sww,0);
          \path[use as bounding box] (0,0) rectangle (\sww,\sz);
        \end{scope}
        \node[font=\bfseries\large] at (sq.center) {$\mathrm{q}$};

        \node[
          draw=none, minimum height=\sz, minimum width=0.4\rectw,
          inner sep=0pt, outer sep=0pt, right=0pt of sq
        ] (rectprmpt) {};
        \begin{scope}[on background layer]
          \fill[#3!10] (rectprmpt.south west) rectangle (rectprmpt.north east);
        \end{scope}
        \draw[#3, line width=1.4\BORDER, dashed]
          (rectprmpt.north west) -- (rectprmpt.north east)
          (rectprmpt.north east) -- (rectprmpt.south east)
          (rectprmpt.south east) -- (rectprmpt.south west);

        \begin{scope}
          \clip (rectprmpt.south west) rectangle (rectprmpt.north east);
        
          \path coordinate (gxW) at ($(rectprmpt.west)!0.19!(rectprmpt.east)$);
          \path coordinate (gxE) at ($(rectprmpt.west)!0.81!(rectprmpt.east)$);
        
          \path coordinate (gy1) at ($(rectprmpt.south)!0.75!(rectprmpt.north)$);
          \fill[#3!60] (gxW|-gy1) circle[radius=1.2pt];
          \draw[#3!60, line width=\BORDER, line cap=round]
               ($(gxW|-gy1)+(1.6pt,0)$) -- ([xshift=-0pt]gxE|-gy1);
        
          \path coordinate (gy2) at ($(rectprmpt.south)!0.50!(rectprmpt.north)$);
          \fill[#3!60] (gxW|-gy2) circle[radius=1.2pt];
          \draw[#3!60, line width=\BORDER, line cap=round]
               ($(gxW|-gy2)+(1.6pt,0)$) -- ([xshift=-3pt]gxE|-gy2);
        
          \path coordinate (gy3) at ($(rectprmpt.south)!0.25!(rectprmpt.north)$);
          \fill[#3!60] (gxW|-gy3) circle[radius=1.2pt];
          \draw[#3!60, line width=\BORDER, line cap=round]
               ($(gxW|-gy3)+(1.6pt,0)$) -- ([xshift=-6pt]gxE|-gy3);
        \end{scope}

        \begin{scope}[local bounding box=rect, xshift=3pt+\sw+0.4\rectw]
          \path[
            draw=#2, line width=1.4\BORDER,
            preaction={fill=#2!10}, pattern color=black!30
          ]
          (0.8\rectw,0) -- (0,0) -- (0,\sz) -- (0.8\rectw,\sz);
        \end{scope}

        \begin{scope}[xshift=3pt+\sw+1.2\rectw, local bounding box=summary]
          \path[
            draw=#3, line width=1.4\BORDER,
            preaction={fill=#3!10}, pattern color=black!30
          ]
          (0,\sz) -- (0.4\rectw,\sz) -- (0.4\rectw,0) -- (0,0);
        \end{scope}

        \begin{scope}[on background layer]
          \fill[#2!10] (rect.south west) rectangle (rect.north east);
          \fill[#3!10] (summary.south west) rectangle (summary.north east);
        \end{scope}
        
        \begin{scope}
          \clip (rect.south west) rectangle (rect.north east);
        
          \def\dia{1.8pt}
        
          \coordinate (gxW) at ($(rect.west)!0.19!(rect.east)$);
          \coordinate (gxE) at ($(rect.west)!0.81!(rect.east)$);
        
          \coordinate (gy1) at ($(rect.south)!0.75!(rect.north)$);
          \coordinate (p1)  at ($(gxW|-gy1)$);
          \fill[#2!60]
            ($(p1)+(-\dia,0)$) -- ($(p1)+(0,\dia)$) -- ($(p1)+(\dia,0)$) -- ($(p1)+(0,-\dia)$) -- cycle;
          \draw[#2!60, line width=\BORDER, line cap=round]
            ($(p1)+(\dia+0.6pt,0)$) -- ([xshift=-2pt]gxE|-gy1);
        
          \coordinate (gy2) at ($(rect.south)!0.50!(rect.north)$);
          \coordinate (p2)  at ($(gxW|-gy2)$);
          \fill[#2!60]
            ($(p2)+(-\dia,0)$) -- ($(p2)+(0,\dia)$) -- ($(p2)+(\dia,0)$) -- ($(p2)+(0,-\dia)$) -- cycle;
          \draw[#2!60, line width=\BORDER, line cap=round]
            ($(p2)+(\dia+0.6pt,0)$) -- ([xshift=-5pt]gxE|-gy2);
        
          \coordinate (gy3) at ($(rect.south)!0.25!(rect.north)$);
          \coordinate (p3)  at ($(gxW|-gy3)$);
          \fill[#2!60]
            ($(p3)+(-\dia,0)$) -- ($(p3)+(0,\dia)$) -- ($(p3)+(\dia,0)$) -- ($(p3)+(0,-\dia)$) -- cycle;
          \draw[#2!60, line width=\BORDER, line cap=round]
            ($(p3)+(\dia+0.6pt,0)$) -- (gxE|-gy3);
        \end{scope}

        \begin{scope}
          \clip (summary.south west) rectangle (summary.north east);
        
          \path coordinate (gxW) at ($(summary.west)!0.18!(summary.east)$);
          \path coordinate (gxE) at ($(summary.west)!0.75!(summary.east)$);
        
          \path coordinate (gy1) at ($(summary.south)!0.75!(summary.north)$);
          \fill[#3!60] (gxW|-gy1) circle[radius=1.2pt];
          \draw[#3!60, line width=\BORDER, line cap=round]
               ($(gxW|-gy1)+(1.6pt,0)$) -- ([xshift=-0pt]gxE|-gy1);
        
          \path coordinate (gy2) at ($(summary.south)!0.50!(summary.north)$);
          \fill[#3!60] (gxW|-gy2) circle[radius=1.2pt];
          \draw[#3!60, line width=\BORDER, line cap=round]
               ($(gxW|-gy2)+(1.6pt,0)$) -- ([xshift=-3pt]gxE|-gy2);
        
          \path coordinate (gy3) at ($(summary.south)!0.25!(summary.north)$);
          \fill[#3!60] (gxW|-gy3) circle[radius=1.2pt];
          \draw[#3!60, line width=\BORDER, line cap=round]
               ($(gxW|-gy3)+(1.6pt,0)$) -- ([xshift=-6pt]gxE|-gy3);
        \end{scope}
        \coordinate (-sqwest)   at (sq.west);
        \coordinate (-rectwest) at (rect.west);
        \coordinate (-recteast) at (rect.east);
        \coordinate (-east)     at (summary.east);
        \coordinate (-north) at (summary.north);
        \coordinate (-rectnorth) at (rect.north);
        \coordinate (-rectsouth) at (rectprmpt.south);
      \end{scope}
    }
  },
  pics/inftythink_iter_n-1/.default={mila-purple-1}{mila-purple-2}
}

\tikzset{
  pics/inftythink_iter_n/.style n args={4}{
    code={
      \begin{scope}[node distance=0pt]
        \begin{scope}[local bounding box=sq]
          \path[draw=none, preaction={fill=#1!10}, pattern color=black!30]
            (\rectr,0) -- (\sww,0) -- (\sww,\sz) -- (\rectr,\sz)
            arc (90:180:\rectr) -- (0,\rectr) arc (180:270:\rectr) -- cycle;
          \draw[#1, dashed, line width=1.4\BORDER]
            (\sww,\sz) -- (\rectr,\sz) arc (90:180:\rectr) -- (0,\rectr)
            arc (180:270:\rectr) -- (\rectr,0) -- (\sww,0);
          \path[use as bounding box] (0,0) rectangle (\sww,\sz);
        \end{scope}
        \node[font=\bfseries\large] at (sq.center) {$\mathrm{q}$};

        \node[
          draw=none, minimum height=\sz, minimum width=0.4\rectw,
          inner sep=0pt, outer sep=0pt, right=0pt of sq
        ] (rectprmpt) {};
        \begin{scope}[on background layer]
          \fill[#3!10] (rectprmpt.south west) rectangle (rectprmpt.north east);
        \end{scope}
        \draw[#3, line width=1.4\BORDER, dashed]
          (rectprmpt.north west) -- (rectprmpt.north east)
          (rectprmpt.north east) -- (rectprmpt.south east)
          (rectprmpt.south east) -- (rectprmpt.south west);

        \begin{scope}
          \clip (rectprmpt.south west) rectangle (rectprmpt.north east);
        
          \path coordinate (gxW) at ($(rectprmpt.west)!0.19!(rectprmpt.east)$);
          \path coordinate (gxE) at ($(rectprmpt.west)!0.81!(rectprmpt.east)$);
        
          \path coordinate (gy1) at ($(rectprmpt.south)!0.75!(rectprmpt.north)$);
          \fill[#3!60] (gxW|-gy1) circle[radius=1.2pt];
          \draw[#3!60, line width=\BORDER, line cap=round]
               ($(gxW|-gy1)+(1.6pt,0)$) -- ([xshift=-0pt]gxE|-gy1);
        
          \path coordinate (gy2) at ($(rectprmpt.south)!0.50!(rectprmpt.north)$);
          \fill[#3!60] (gxW|-gy2) circle[radius=1.2pt];
          \draw[#3!60, line width=\BORDER, line cap=round]
               ($(gxW|-gy2)+(1.6pt,0)$) -- ([xshift=-3pt]gxE|-gy2);
        
          \path coordinate (gy3) at ($(rectprmpt.south)!0.25!(rectprmpt.north)$);
          \fill[#3!60] (gxW|-gy3) circle[radius=1.2pt];
          \draw[#3!60, line width=\BORDER, line cap=round]
               ($(gxW|-gy3)+(1.6pt,0)$) -- ([xshift=-6pt]gxE|-gy3);
        \end{scope}

        \begin{scope}[local bounding box=rect, xshift=3pt+\sw+0.4\rectw]
          \path[
            draw=#2, line width=1.4\BORDER,
            preaction={fill=#2!10}, pattern color=black!30
          ]
          (0.8\rectw,0) -- (0,0) -- (0,\sz) -- (0.8\rectw,\sz);
        \end{scope}

        \begin{scope}[xshift=3pt+\sw+1.2\rectw, local bounding box=conclusion]
          \path[
            draw=#4, line width=1.4\BORDER,
            preaction={fill=#4!10}, pattern color=black!30
          ]
          (0,\sz) -- (\rectwconclu,\sz) 
          arc (90:0:\rectr) -- (\rectwconclu+\rectr,\rectr) arc (0:-90:\rectr) -- (\rectwconclu,0) -- (0,0);
        \end{scope}
        
        \begin{scope}[on background layer]
          \fill[#2!10] (rect.south west) rectangle (rect.north east);
        \end{scope}
        
        \begin{scope}
          \clip (rect.south west) rectangle (rect.north east);
        
          \def\dia{1.8pt}
        
          \coordinate (gxW) at ($(rect.west)!0.19!(rect.east)$);
          \coordinate (gxE) at ($(rect.west)!0.81!(rect.east)$);
        
          \coordinate (gy1) at ($(rect.south)!0.75!(rect.north)$);
          \coordinate (p1)  at ($(gxW|-gy1)$);
          \fill[#2!60]
            ($(p1)+(-\dia,0)$) -- ($(p1)+(0,\dia)$) -- ($(p1)+(\dia,0)$) -- ($(p1)+(0,-\dia)$) -- cycle;
          \draw[#2!60, line width=\BORDER, line cap=round]
            ($(p1)+(\dia+0.6pt,0)$) -- ([xshift=-2pt]gxE|-gy1);
        
          \coordinate (gy2) at ($(rect.south)!0.50!(rect.north)$);
          \coordinate (p2)  at ($(gxW|-gy2)$);
          \fill[#2!60]
            ($(p2)+(-\dia,0)$) -- ($(p2)+(0,\dia)$) -- ($(p2)+(\dia,0)$) -- ($(p2)+(0,-\dia)$) -- cycle;
          \draw[#2!60, line width=\BORDER, line cap=round]
            ($(p2)+(\dia+0.6pt,0)$) -- ([xshift=-5pt]gxE|-gy2);
        
          \coordinate (gy3) at ($(rect.south)!0.25!(rect.north)$);
          \coordinate (p3)  at ($(gxW|-gy3)$);
          \fill[#2!60]
            ($(p3)+(-\dia,0)$) -- ($(p3)+(0,\dia)$) -- ($(p3)+(\dia,0)$) -- ($(p3)+(0,-\dia)$) -- cycle;
          \draw[#2!60, line width=\BORDER, line cap=round]
            ($(p3)+(\dia+0.6pt,0)$) -- (gxE|-gy3);
        \end{scope}

        \begin{scope}
          \clip (conclusion.south west) rectangle (conclusion.north east);
        
          \def\triH{1.2pt}   %
          \def\triL{1.8pt}   %
        
          \coordinate (gxW) at ($(conclusion.west)!0.18!(conclusion.east)$);
          \coordinate (gxE) at ($(conclusion.west)!0.73!(conclusion.east)$);
        
          \coordinate (gy1) at ($(conclusion.south)!0.75!(conclusion.north)$);
          \coordinate (p1)  at ($(gxW|-gy1)$);
          \fill[#4!60]
            ($(p1)+(-\triL,-\triH)$) --
            ($(p1)+(-\triL, \triH)$) --
            ($(p1)+(\triL,0)$) -- cycle;
          \draw[#4!60, line width=\BORDER, line cap=round]
            ($(p1)+(\triL+0.6pt,0)$) -- ([xshift=-3pt]gxE|-gy1);
        
          \coordinate (gy2) at ($(conclusion.south)!0.50!(conclusion.north)$);
          \coordinate (p2)  at ($(gxW|-gy2)$);
          \fill[#4!60]
            ($(p2)+(-\triL,-\triH)$) --
            ($(p2)+(-\triL, \triH)$) --
            ($(p2)+(\triL,0)$) -- cycle;
          \draw[#4!60, line width=\BORDER, line cap=round]
            ($(p2)+(\triL+0.6pt,0)$) -- ([xshift=-5pt]gxE|-gy2);
        
          \coordinate (gy3) at ($(conclusion.south)!0.25!(conclusion.north)$);
          \coordinate (p3)  at ($(gxW|-gy3)$);
          \fill[#4!60]
            ($(p3)+(-\triL,-\triH)$) --
            ($(p3)+(-\triL, \triH)$) --
            ($(p3)+(\triL,0)$) -- cycle;
          \draw[#4!60, line width=\BORDER, line cap=round]
            ($(p3)+(\triL+0.6pt,0)$) -- (gxE|-gy3);
        \end{scope}

        \coordinate (-sqwest)   at (sq.west);
        \coordinate (-rectwest) at (rect.west);
        \coordinate (-recteast) at (rect.east);
        \coordinate (-east)     at (conclusion.east);
        \coordinate (-north) at (conclusion.north);
        \coordinate (-rectnorth) at (rect.north);
        \coordinate (-rectsouth) at (rectprmpt.south);
      \end{scope}
    }
  },
  pics/inftythink_iter_n/.default={mila-purple-1}{mila-purple-2}
}

\tikzset{
  pics/segment/.style n args={2}{
    code={
      \begin{scope}[node distance=0pt]
        \begin{scope}[local bounding box=sq]
          \path[
            draw=delethink-purple, line width=1.4\BORDER, dashed,
            preaction={fill=delethink-purple!5}, pattern color=black!30
          ]
          (\rectr,0) -- (\sw,0) -- (\sw,\sz) -- (\rectr,\sz)
          arc (90:180:\rectr) -- (0,\rectr) arc (180:270:\rectr) -- cycle;
        \end{scope}
        \node[font=\bfseries\large] at (sq.center) {$\mathrm{q}$};

        \node[
          draw=delethink-purple, line width=1.4\BORDER,
          minimum height=\sz, minimum width=\rectw, inner sep=0pt, right=3pt of sq
        ] (rect) {};

        \draw[dashed, line cap=round,
      shorten >=-3pt, shorten <=-3pt](rect.north) -- (rect.south);

        \begin{scope}[on background layer]
          \fill[#1] (rect.south west) rectangle (rect.north);
          \fill[#2] (rect.south) rectangle (rect.north east);
        \end{scope}

        \begin{scope}
          \clip (rect.south west) rectangle (rect.north east);
        
          \def\dia{1.8pt}
        
          \coordinate (gxW) at ($(rect.west)!0.12!(rect.east)$);
          \coordinate (gxE) at ($(rect.west)!0.42!(rect.east)$);
        
          \coordinate (gy1) at ($(rect.south)!0.75!(rect.north)$);
          \coordinate (p1)  at ($(gxW|-gy1)$);
          \fill[delethink-purple!40]
            ($(p1)+(-\dia,0)$) -- ($(p1)+(0,\dia)$) -- ($(p1)+(\dia,0)$) -- ($(p1)+(0,-\dia)$) -- cycle;
          \draw[delethink-purple!40, line width=\BORDER, line cap=round]
            ($(p1)+(\dia+0.6pt,0)$) -- ([xshift=-2pt]gxE|-gy1);
        
          \coordinate (gy2) at ($(rect.south)!0.50!(rect.north)$);
          \coordinate (p2)  at ($(gxW|-gy2)$);
          \fill[delethink-purple!40]
            ($(p2)+(-\dia,0)$) -- ($(p2)+(0,\dia)$) -- ($(p2)+(\dia,0)$) -- ($(p2)+(0,-\dia)$) -- cycle;
          \draw[delethink-purple!40, line width=\BORDER, line cap=round]
            ($(p2)+(\dia+0.6pt,0)$) -- ([xshift=-5pt]gxE|-gy2);
        
          \coordinate (gy3) at ($(rect.south)!0.25!(rect.north)$);
          \coordinate (p3)  at ($(gxW|-gy3)$);
          \fill[delethink-purple!40]
            ($(p3)+(-\dia,0)$) -- ($(p3)+(0,\dia)$) -- ($(p3)+(\dia,0)$) -- ($(p3)+(0,-\dia)$) -- cycle;
          \draw[delethink-purple!40, line width=\BORDER, line cap=round]
            ($(p3)+(\dia+0.6pt,0)$) -- (gxE|-gy3);
        \end{scope}

        \begin{scope}
          \clip (rect.south west) rectangle (rect.north east);
        
          \def\dia{1.8pt}
        
          \coordinate (gxW) at ($(rect.west)!0.585!(rect.east)$);
          \coordinate (gxE) at ($(rect.west)!0.89!(rect.east)$);
        
          \coordinate (gy1) at ($(rect.south)!0.75!(rect.north)$);
          \coordinate (p1)  at ($(gxW|-gy1)$);
          \fill[delethink-purple!40]
            ($(p1)+(-\dia,0)$) -- ($(p1)+(0,\dia)$) -- ($(p1)+(\dia,0)$) -- ($(p1)+(0,-\dia)$) -- cycle;
          \draw[delethink-purple!40, line width=\BORDER, line cap=round]
            ($(p1)+(\dia+0.6pt,0)$) -- ([xshift=-7pt]gxE|-gy1);
        
          \coordinate (gy2) at ($(rect.south)!0.50!(rect.north)$);
          \coordinate (p2)  at ($(gxW|-gy2)$);
          \fill[delethink-purple!40]
            ($(p2)+(-\dia,0)$) -- ($(p2)+(0,\dia)$) -- ($(p2)+(\dia,0)$) -- ($(p2)+(0,-\dia)$) -- cycle;
          \draw[delethink-purple!40, line width=\BORDER, line cap=round]
            ($(p2)+(\dia+0.6pt,0)$) -- ([xshift=-1pt]gxE|-gy2);
        
          \coordinate (gy3) at ($(rect.south)!0.25!(rect.north)$);
          \coordinate (p3)  at ($(gxW|-gy3)$);
          \fill[delethink-purple!40]
            ($(p3)+(-\dia,0)$) -- ($(p3)+(0,\dia)$) -- ($(p3)+(\dia,0)$) -- ($(p3)+(0,-\dia)$) -- cycle;
          \draw[delethink-purple!40, line width=\BORDER, line cap=round]
            ($(p3)+(\dia+0.6pt,0)$) -- ([xshift=-5pt]gxE|-gy3);
        \end{scope}

        \coordinate (-sqwest)   at (sq.west);
        \coordinate (-rectwest) at (rect.west);
        \coordinate (-recteast) at (rect.east);
        \coordinate (-east)     at (rect.east);
        \coordinate (-rectnorth) at ([xshift=0.25\rectw]rect.north);
        \coordinate (-rectsouth) at (rect.south);
      \end{scope}
    }
  },
  pics/segment/.default={mila-purple-1}{mila-purple-2}
}

\tikzset{
  pics/segmentlongcot/.style n args={3}{
    code={
      \begin{scope}[node distance=0pt]
        \begin{scope}[local bounding box=sq]
          \path[
            draw=#1, line width=1.4\BORDER, dashed,
            preaction={fill=#1!10}, pattern color=black!30
          ]
          (\rectr,0) -- (\sw,0) -- (\sw,\sz) -- (\rectr,\sz)
          arc (90:180:\rectr) -- (0,\rectr) arc (180:270:\rectr) -- cycle;
        \end{scope}
        \node[font=\bfseries\large] at (sq.center) {$\mathrm{q}$};

        \begin{scope}[local bounding box=rect, xshift=3pt+\sw]
          \path[
            draw=#2, line width=1.4\BORDER,
            preaction={fill=#2!10}, pattern color=black!30
          ]
          (\rectwlongcot,0) -- (0,0) -- (0,\sz) -- (\rectwlongcot,\sz);
        \end{scope}

        \begin{scope}[xshift=3pt+\sw+\rectwlongcot, local bounding box=conclusion]
          \path[
            draw=#3, line width=1.4\BORDER,
            preaction={fill=#3!10}, pattern color=black!30
          ]
          (0,\sz) -- (\rectwconclu,\sz) 
          arc (90:0:\rectr) -- (\rectwconclu+\rectr,\rectr) arc (0:-90:\rectr) -- (\rectwconclu,0) -- (0,0);
        \end{scope}

        \begin{scope}[on background layer]
          \fill[#2!10] (rect.south west) rectangle (rect.north east);
        \end{scope}

        \begin{scope}
          \clip (rect.south west) rectangle (rect.north east);
        
          \def\dia{1.8pt}
        
          \coordinate (gxW) at ($(rect.west)!0.019!(rect.east)$);
          \coordinate (gxE) at ($(rect.west)!0.065!(rect.east)$);
        
          \coordinate (gy1) at ($(rect.south)!0.75!(rect.north)$);
          \coordinate (p1)  at ($(gxW|-gy1)$);
          \fill[delethink-blue!60]
            ($(p1)+(-\dia,0)$) -- ($(p1)+(0,\dia)$) -- ($(p1)+(\dia,0)$) -- ($(p1)+(0,-\dia)$) -- cycle;
          \draw[delethink-blue!60, line width=\BORDER, line cap=round]
            ($(p1)+(\dia+0.6pt,0)$) -- ([xshift=-2pt]gxE|-gy1);
        
          \coordinate (gy2) at ($(rect.south)!0.50!(rect.north)$);
          \coordinate (p2)  at ($(gxW|-gy2)$);
          \fill[delethink-blue!60]
            ($(p2)+(-\dia,0)$) -- ($(p2)+(0,\dia)$) -- ($(p2)+(\dia,0)$) -- ($(p2)+(0,-\dia)$) -- cycle;
          \draw[delethink-blue!60, line width=\BORDER, line cap=round]
            ($(p2)+(\dia+0.6pt,0)$) -- ([xshift=-5pt]gxE|-gy2);
        
          \coordinate (gy3) at ($(rect.south)!0.25!(rect.north)$);
          \coordinate (p3)  at ($(gxW|-gy3)$);
          \fill[delethink-blue!60]
            ($(p3)+(-\dia,0)$) -- ($(p3)+(0,\dia)$) -- ($(p3)+(\dia,0)$) -- ($(p3)+(0,-\dia)$) -- cycle;
          \draw[delethink-blue!60, line width=\BORDER, line cap=round]
            ($(p3)+(\dia+0.6pt,0)$) -- ([xshift=-0pt]gxE|-gy3);
        \end{scope}

        \begin{scope}
          \clip (rect.south west) rectangle (rect.north east);
        
          \def\dia{1.8pt}
        
          \coordinate (gxW) at ($(rect.west)!0.085!(rect.east)$);
          \coordinate (gxE) at ($(rect.west)!0.130!(rect.east)$);
        
          \coordinate (gy1) at ($(rect.south)!0.75!(rect.north)$);
          \coordinate (p1)  at ($(gxW|-gy1)$);
          \fill[delethink-blue!50]
            ($(p1)+(-\dia,0)$) -- ($(p1)+(0,\dia)$) -- ($(p1)+(\dia,0)$) -- ($(p1)+(0,-\dia)$) -- cycle;
          \draw[delethink-blue!50, line width=\BORDER, line cap=round]
            ($(p1)+(\dia+0.6pt,0)$) -- ([xshift=-6pt]gxE|-gy1);
        
          \coordinate (gy2) at ($(rect.south)!0.50!(rect.north)$);
          \coordinate (p2)  at ($(gxW|-gy2)$);
          \fill[delethink-blue!50]
            ($(p2)+(-\dia,0)$) -- ($(p2)+(0,\dia)$) -- ($(p2)+(\dia,0)$) -- ($(p2)+(0,-\dia)$) -- cycle;
          \draw[delethink-blue!50, line width=\BORDER, line cap=round]
            ($(p2)+(\dia+0.6pt,0)$) -- ([xshift=-0pt]gxE|-gy2);
        
          \coordinate (gy3) at ($(rect.south)!0.25!(rect.north)$);
          \coordinate (p3)  at ($(gxW|-gy3)$);
          \fill[delethink-blue!50]
            ($(p3)+(-\dia,0)$) -- ($(p3)+(0,\dia)$) -- ($(p3)+(\dia,0)$) -- ($(p3)+(0,-\dia)$) -- cycle;
          \draw[delethink-blue!50, line width=\BORDER, line cap=round]
            ($(p3)+(\dia+0.6pt,0)$) -- ([xshift=-4pt]gxE|-gy3);
        \end{scope}

        \begin{scope}
          \clip (rect.south west) rectangle (rect.north east);
        
          \def\dia{1.8pt}
        
          \coordinate (gxW) at ($(rect.west)!0.15!(rect.east)$);
          \coordinate (gxE) at ($(rect.west)!0.19!(rect.east)$);
        
          \coordinate (gy1) at ($(rect.south)!0.75!(rect.north)$);
          \coordinate (p1)  at ($(gxW|-gy1)$);
          \fill[delethink-blue!40]
            ($(p1)+(-\dia,0)$) -- ($(p1)+(0,\dia)$) -- ($(p1)+(\dia,0)$) -- ($(p1)+(0,-\dia)$) -- cycle;
          \draw[delethink-blue!40, line width=\BORDER, line cap=round]
            ($(p1)+(\dia+0.6pt,0)$) -- ([xshift=-0pt]gxE|-gy1);
        
          \coordinate (gy2) at ($(rect.south)!0.50!(rect.north)$);
          \coordinate (p2)  at ($(gxW|-gy2)$);
          \fill[delethink-blue!40]
            ($(p2)+(-\dia,0)$) -- ($(p2)+(0,\dia)$) -- ($(p2)+(\dia,0)$) -- ($(p2)+(0,-\dia)$) -- cycle;
          \draw[delethink-blue!40, line width=\BORDER, line cap=round]
            ($(p2)+(\dia+0.6pt,0)$) -- ([xshift=-9pt]gxE|-gy2);
        
          \coordinate (gy3) at ($(rect.south)!0.25!(rect.north)$);
          \coordinate (p3)  at ($(gxW|-gy3)$);
          \fill[delethink-blue!40]
            ($(p3)+(-\dia,0)$) -- ($(p3)+(0,\dia)$) -- ($(p3)+(\dia,0)$) -- ($(p3)+(0,-\dia)$) -- cycle;
          \draw[delethink-blue!40, line width=\BORDER, line cap=round]
            ($(p3)+(\dia+0.6pt,0)$) -- ([xshift=-6pt]gxE|-gy3);
        \end{scope}

        \begin{scope}
          \clip (rect.south west) rectangle (rect.north east);
        
          \def\dia{1.8pt}
        
          \coordinate (gxW) at ($(rect.west)!0.21!(rect.east)$);
          \coordinate (gxE) at ($(rect.west)!0.26!(rect.east)$);
        
          \coordinate (gy1) at ($(rect.south)!0.75!(rect.north)$);
          \coordinate (p1)  at ($(gxW|-gy1)$);
          \fill[delethink-blue!30]
            ($(p1)+(-\dia,0)$) -- ($(p1)+(0,\dia)$) -- ($(p1)+(\dia,0)$) -- ($(p1)+(0,-\dia)$) -- cycle;
          \draw[delethink-blue!30, line width=\BORDER, line cap=round]
            ($(p1)+(\dia+0.6pt,0)$) -- ([xshift=-5pt]gxE|-gy1);
        
          \coordinate (gy2) at ($(rect.south)!0.50!(rect.north)$);
          \coordinate (p2)  at ($(gxW|-gy2)$);
          \fill[delethink-blue!30]
            ($(p2)+(-\dia,0)$) -- ($(p2)+(0,\dia)$) -- ($(p2)+(\dia,0)$) -- ($(p2)+(0,-\dia)$) -- cycle;
          \draw[delethink-blue!30, line width=\BORDER, line cap=round]
            ($(p2)+(\dia+0.6pt,0)$) -- ([xshift=-6pt]gxE|-gy2);
        
          \coordinate (gy3) at ($(rect.south)!0.25!(rect.north)$);
          \coordinate (p3)  at ($(gxW|-gy3)$);
          \fill[delethink-blue!30]
            ($(p3)+(-\dia,0)$) -- ($(p3)+(0,\dia)$) -- ($(p3)+(\dia,0)$) -- ($(p3)+(0,-\dia)$) -- cycle;
          \draw[delethink-blue!30, line width=\BORDER, line cap=round]
            ($(p3)+(\dia+0.6pt,0)$) -- ([xshift=-0pt]gxE|-gy3);
        \end{scope}

        \begin{scope}
          \clip (conclusion.south west) rectangle (conclusion.north east);
        
          \def\triH{1.2pt}   %
          \def\triL{1.8pt}   %
        
          \coordinate (gxW) at ($(conclusion.west)!0.18!(conclusion.east)$);
          \coordinate (gxE) at ($(conclusion.west)!0.73!(conclusion.east)$);
        
          \coordinate (gy1) at ($(conclusion.south)!0.75!(conclusion.north)$);
          \coordinate (p1)  at ($(gxW|-gy1)$);
          \fill[#3!60]
            ($(p1)+(-\triL,-\triH)$) --
            ($(p1)+(-\triL, \triH)$) --
            ($(p1)+(\triL,0)$) -- cycle;
          \draw[#3!60, line width=\BORDER, line cap=round]
            ($(p1)+(\triL+0.6pt,0)$) -- ([xshift=-3pt]gxE|-gy1);
        
          \coordinate (gy2) at ($(conclusion.south)!0.50!(conclusion.north)$);
          \coordinate (p2)  at ($(gxW|-gy2)$);
          \fill[#3!60]
            ($(p2)+(-\triL,-\triH)$) --
            ($(p2)+(-\triL, \triH)$) --
            ($(p2)+(\triL,0)$) -- cycle;
          \draw[#3!60, line width=\BORDER, line cap=round]
            ($(p2)+(\triL+0.6pt,0)$) -- ([xshift=-5pt]gxE|-gy2);
        
          \coordinate (gy3) at ($(conclusion.south)!0.25!(conclusion.north)$);
          \coordinate (p3)  at ($(gxW|-gy3)$);
          \fill[#3!60]
            ($(p3)+(-\triL,-\triH)$) --
            ($(p3)+(-\triL, \triH)$) --
            ($(p3)+(\triL,0)$) -- cycle;
          \draw[#3!60, line width=\BORDER, line cap=round]
            ($(p3)+(\triL+0.6pt,0)$) -- (gxE|-gy3);
        \end{scope}

        \coordinate (-sqwest)   at (sq.west);
        \coordinate (-rectwest) at (rect.west);
        \coordinate (-recteast) at (rect.east);
        \coordinate (-east)     at (conclusion.east);
        \coordinate (-rectnorth) at ([xshift=0.25\rectw]rect.north);
        \coordinate (-rectsouth) at (rect.south);
      \end{scope}
    }
  },
  pics/segmentlongcot/.default={mila-purple-1}{mila-purple}
}

\tikzset{
  pics/segmentintermediate/.style n args={3}{
    code={
      \begin{scope}[node distance=0pt]
        \begin{scope}[local bounding box=sq]
          \path[draw=none, preaction={fill=delethink-purple!5}, pattern color=black!30]
            (\rectr,0) -- (\sww,0) -- (\sww,\sz) -- (\rectr,\sz)
            arc (90:180:\rectr) -- (0,\rectr) arc (180:270:\rectr) -- cycle;
          \draw[delethink-purple, dashed, line width=1.4\BORDER]
            (\sww,\sz) -- (\rectr,\sz) arc (90:180:\rectr) -- (0,\rectr)
            arc (180:270:\rectr) -- (\rectr,0) -- (\sww,0);
          \path[use as bounding box] (0,0) rectangle (\sww,\sz);
        \end{scope}
        \node[font=\bfseries\large] at (sq.center) {$\mathrm{q}$};

        \node[
          draw=none, minimum height=\sz, minimum width=0.42\rectw,
          inner sep=0pt, outer sep=0pt, right=0pt of sq
        ] (rectprmpt) {};
        \begin{scope}[on background layer]
          \fill[#1] (rectprmpt.south west) rectangle (rectprmpt.north east);
        \end{scope}
        \draw[delethink-purple, line width=1.4\BORDER, dashed]
          (rectprmpt.north west) -- (rectprmpt.north east)
          (rectprmpt.north east) -- (rectprmpt.south east)
          (rectprmpt.south east) -- (rectprmpt.south west);

        \node[
          draw=#3, line width=1.4\BORDER,
          minimum height=\sz, minimum width=0.5\rectw,
          inner sep=0pt, outer sep=0pt, right=4pt of rectprmpt
        ] (rect) {};
        \begin{scope}[on background layer]
          \fill[#2] (rect.south west) rectangle (rect.north east);
        \end{scope}

        \begin{scope}
          \clip (rect.south west) rectangle (rect.north east);
        
          \path coordinate (gxW) at ($(rect.west)!0.18!(rect.east)$);
          \path coordinate (gxE) at ($(rect.west)!0.80!(rect.east)$);
        
          \path coordinate (gy1) at ($(rect.south)!0.75!(rect.north)$);
          \fill[white!60] (gxW|-gy1) circle[radius=1.2pt];
          \draw[white!60, line width=\BORDER, line cap=round]
               ($(gxW|-gy1)+(1.6pt,0)$) -- (gxE|-gy1);
        
          \path coordinate (gy2) at ($(rect.south)!0.50!(rect.north)$);
          \fill[white!60] (gxW|-gy2) circle[radius=1.2pt];
          \draw[white!60, line width=\BORDER, line cap=round]
               ($(gxW|-gy2)+(1.6pt,0)$) -- (gxE|-gy2);
        
          \path coordinate (gy3) at ($(rect.south)!0.25!(rect.north)$);
          \fill[white!60] (gxW|-gy3) circle[radius=1.2pt];
          \draw[white!60, line width=\BORDER, line cap=round]
               ($(gxW|-gy3)+(1.6pt,0)$) -- (gxE|-gy3);
        \end{scope}

        \coordinate (-sqwest)   at (sq.west);
        \coordinate (-rectwest) at (rect.west);
        \coordinate (-recteast) at (rect.east);
        \coordinate (-east)     at (rect.east);
        \coordinate (-rectnorth) at (rect.north);
        \coordinate (-rectsouth) at (rectprmpt.south);
      \end{scope}
    }
  },
  pics/segmentintermediate/.default={mila-purple-1}{mila-purple-2}{mila-purple}
}

\tikzset{
  pics/segmentintermediatecircle/.style n args={3}{
    code={
      \begin{scope}[node distance=0pt]
        \begin{scope}[local bounding box=sq]
          \path[draw=none, preaction={fill=delethink-purple!5}, pattern color=black!30]
            (\rectr,0) -- (\sww,0) -- (\sww,\sz) -- (\rectr,\sz)
            arc (90:180:\rectr) -- (0,\rectr) arc (180:270:\rectr) -- cycle;
          \draw[delethink-purple, dashed, line width=1.4\BORDER]
            (\sww,\sz) -- (\rectr,\sz) arc (90:180:\rectr) -- (0,\rectr)
            arc (180:270:\rectr) -- (\rectr,0) -- (\sww,0);
          \path[use as bounding box] (0,0) rectangle (\sww,\sz);
        \end{scope}
        \node[font=\bfseries\large] at (sq.center) {$\mathrm{q}$};

        \node[
          draw=none, minimum height=\sz, minimum width=0.42\rectw,
          inner sep=0pt, outer sep=0pt, right=0pt of sq
        ] (rectprmpt) {};
        \begin{scope}[on background layer]
          \fill[#1] (rectprmpt.south west) rectangle (rectprmpt.north east);
        \end{scope}
        \draw[delethink-purple, line width=1.4\BORDER, dashed]
          (rectprmpt.north west) -- (rectprmpt.north east)
          (rectprmpt.north east) -- (rectprmpt.south east)
          (rectprmpt.south east) -- (rectprmpt.south west);

        \begin{scope}
          \clip (rectprmpt.south west) rectangle (rectprmpt.north east);
        
          \def\dia{1.8pt}
        
          \coordinate (gxW) at ($(rectprmpt.west)!0.19!(rectprmpt.east)$);
          \coordinate (gxE) at ($(rectprmpt.west)!0.82!(rectprmpt.east)$);
        
          \coordinate (gy1) at ($(rectprmpt.south)!0.75!(rectprmpt.north)$);
          \coordinate (p1)  at ($(gxW|-gy1)$);
          \fill[delethink-purple!40]
            ($(p1)+(-\dia,0)$) -- ($(p1)+(0,\dia)$) -- ($(p1)+(\dia,0)$) -- ($(p1)+(0,-\dia)$) -- cycle;
          \draw[delethink-purple!40, line width=\BORDER, line cap=round]
            ($(p1)+(\dia+0.6pt,0)$) -- ([xshift=-6pt]gxE|-gy1);
        
          \coordinate (gy2) at ($(rectprmpt.south)!0.50!(rectprmpt.north)$);
          \coordinate (p2)  at ($(gxW|-gy2)$);
          \fill[delethink-purple!40]
            ($(p2)+(-\dia,0)$) -- ($(p2)+(0,\dia)$) -- ($(p2)+(\dia,0)$) -- ($(p2)+(0,-\dia)$) -- cycle;
          \draw[delethink-purple!40, line width=\BORDER, line cap=round]
            ($(p2)+(\dia+0.6pt,0)$) -- ([xshift=-0pt]gxE|-gy2);
        
          \coordinate (gy3) at ($(rectprmpt.south)!0.25!(rectprmpt.north)$);
          \coordinate (p3)  at ($(gxW|-gy3)$);
          \fill[delethink-purple!40]
            ($(p3)+(-\dia,0)$) -- ($(p3)+(0,\dia)$) -- ($(p3)+(\dia,0)$) -- ($(p3)+(0,-\dia)$) -- cycle;
          \draw[delethink-purple!40, line width=\BORDER, line cap=round]
            ($(p3)+(\dia+0.6pt,0)$) -- ([xshift=-4pt]gxE|-gy3);
        \end{scope}

        \node[
          draw=#3, line width=1.4\BORDER,
          minimum height=\sz, minimum width=0.5\rectw,
          inner sep=0pt, outer sep=0pt, right=4pt of rectprmpt
        ] (rect) {};
        \begin{scope}[on background layer]
          \fill[#2] (rect.south west) rectangle (rect.north east);
        \end{scope}

        \begin{scope}
          \clip (rect.south west) rectangle (rect.north east);
        
          \path coordinate (gxW) at ($(rect.west)!0.18!(rect.east)$);
          \path coordinate (gxE) at ($(rect.west)!0.75!(rect.east)$);
        
          \path coordinate (gy1) at ($(rect.south)!0.75!(rect.north)$);
          \fill[white!60] (gxW|-gy1) circle[radius=1.2pt];
          \draw[white!60, line width=\BORDER, line cap=round]
               ($(gxW|-gy1)+(1.6pt,0)$) -- ([xshift=-0pt]gxE|-gy1);
        
          \path coordinate (gy2) at ($(rect.south)!0.50!(rect.north)$);
          \fill[white!60] (gxW|-gy2) circle[radius=1.2pt];
          \draw[white!60, line width=\BORDER, line cap=round]
               ($(gxW|-gy2)+(1.6pt,0)$) -- ([xshift=-3pt]gxE|-gy2);
        
          \path coordinate (gy3) at ($(rect.south)!0.25!(rect.north)$);
          \fill[white!60] (gxW|-gy3) circle[radius=1.2pt];
          \draw[white!60, line width=\BORDER, line cap=round]
               ($(gxW|-gy3)+(1.6pt,0)$) -- ([xshift=-6pt]gxE|-gy3);
        \end{scope}

        \coordinate (-sqwest)   at (sq.west);
        \coordinate (-rectwest) at (rect.west);
        \coordinate (-recteast) at (rect.east);
        \coordinate (-east)     at (rect.east);
        \coordinate (-rectnorth) at (rect.north);
        \coordinate (-rectsouth) at (rectprmpt.south);
      \end{scope}
    }
  },
  pics/segmentintermediatecircle/.default={mila-purple-1}{mila-purple-2}{mila-purple}
}

\tikzset{
  pics/segmentintermediatesquare/.style n args={3}{
    code={
      \begin{scope}[node distance=0pt]
        \begin{scope}[local bounding box=sq]
          \path[draw=none, preaction={fill=delethink-purple!5}, pattern color=black!30]
            (\rectr,0) -- (\sww,0) -- (\sww,\sz) -- (\rectr,\sz)
            arc (90:180:\rectr) -- (0,\rectr) arc (180:270:\rectr) -- cycle;
          \draw[delethink-purple, dashed, line width=1.4\BORDER]
            (\sww,\sz) -- (\rectr,\sz) arc (90:180:\rectr) -- (0,\rectr)
            arc (180:270:\rectr) -- (\rectr,0) -- (\sww,0);
          \path[use as bounding box] (0,0) rectangle (\sww,\sz);
        \end{scope}
        \node[font=\bfseries\large] at (sq.center) {$\mathrm{q}$};

        \node[
          draw=none, minimum height=\sz, minimum width=0.42\rectw,
          inner sep=0pt, outer sep=0pt, right=0pt of sq
        ] (rectprmpt) {};
        \begin{scope}[on background layer]
          \fill[#1] (rectprmpt.south west) rectangle (rectprmpt.north east);
        \end{scope}
        \draw[delethink-purple, line width=1.4\BORDER, dashed]
          (rectprmpt.north west) -- (rectprmpt.north east)
          (rectprmpt.north east) -- (rectprmpt.south east)
          (rectprmpt.south east) -- (rectprmpt.south west);

        \begin{scope}
          \clip (rectprmpt.south west) rectangle (rectprmpt.north east);
        
          \def\triH{1.2pt}   %
          \def\triL{1.8
          pt}   %
        
          \coordinate (gxW) at ($(rectprmpt.west)!0.19!(rectprmpt.east)$);
          \coordinate (gxE) at ($(rectprmpt.west)!0.82!(rectprmpt.east)$);
        
          \coordinate (gy1) at ($(rectprmpt.south)!0.75!(rectprmpt.north)$);
          \coordinate (p1)  at ($(gxW|-gy1)$);
          \fill[white]
            ($(p1)+(-\triL,-\triH)$) --
            ($(p1)+(-\triL, \triH)$) --
            ($(p1)+(\triL,0)$) -- cycle;
          \draw[white, line width=\BORDER, line cap=round]
            ($(p1)+(\triL+0.6pt,0)$) -- ([xshift=-3pt]gxE|-gy1);
        
          \coordinate (gy2) at ($(rectprmpt.south)!0.50!(rectprmpt.north)$);
          \coordinate (p2)  at ($(gxW|-gy2)$);
          \fill[white]
            ($(p2)+(-\triL,-\triH)$) --
            ($(p2)+(-\triL, \triH)$) --
            ($(p2)+(\triL,0)$) -- cycle;
          \draw[white, line width=\BORDER, line cap=round]
            ($(p2)+(\triL+0.6pt,0)$) -- ([xshift=-5pt]gxE|-gy2);
        
          \coordinate (gy3) at ($(rectprmpt.south)!0.25!(rectprmpt.north)$);
          \coordinate (p3)  at ($(gxW|-gy3)$);
          \fill[white]
            ($(p3)+(-\triL,-\triH)$) --
            ($(p3)+(-\triL, \triH)$) --
            ($(p3)+(\triL,0)$) -- cycle;
          \draw[white, line width=\BORDER, line cap=round]
            ($(p3)+(\triL+0.6pt,0)$) -- (gxE|-gy3);
        \end{scope}

        \node[
          draw=#3, line width=1.4\BORDER,
          minimum height=\sz, minimum width=0.5\rectw,
          inner sep=0pt, outer sep=0pt, right=4pt of rectprmpt
        ] (rect) {};
        \begin{scope}[on background layer]
          \fill[#2] (rect.south west) rectangle (rect.north east);
        \end{scope}

        \begin{scope}
          \clip (rect.south west) rectangle (rect.north east);
        
          \def\sq{2.2pt} %
          \path coordinate (gxW) at ($(rect.west)!0.19!(rect.east)$);
          \path coordinate (gxE) at ($(rect.west)!0.75!(rect.east)$);
        
          \path coordinate (gy1) at ($(rect.south)!0.75!(rect.north)$);
          \fill[white] ($(gxW|-gy1)+(-.5*\sq,-.5*\sq)$) rectangle ++(\sq,\sq);
          \draw[white, line width=\BORDER, line cap=round]
               ($(gxW|-gy1)+(.5*\sq+0.6pt,0)$) -- ([xshift=-8pt]gxE|-gy1);
        
          \path coordinate (gy2) at ($(rect.south)!0.50!(rect.north)$);
          \fill[white] ($(gxW|-gy2)+(-.5*\sq,-.5*\sq)$) rectangle ++(\sq,\sq);
          \draw[white, line width=\BORDER, line cap=round]
               ($(gxW|-gy2)+(.5*\sq+0.6pt,0)$) -- ([xshift=-5pt]gxE|-gy2);
        
          \path coordinate (gy3) at ($(rect.south)!0.25!(rect.north)$);
          \fill[white] ($(gxW|-gy3)+(-.5*\sq,-.5*\sq)$) rectangle ++(\sq,\sq);
          \draw[white, line width=\BORDER, line cap=round]
               ($(gxW|-gy3)+(.5*\sq+0.6pt,0)$) -- (gxE|-gy3);
        \end{scope}

        \coordinate (-sqwest)   at (sq.west);
        \coordinate (-rectwest) at (rect.west);
        \coordinate (-recteast) at (rect.east);
        \coordinate (-east)     at (rect.east);
        \coordinate (-rectnorth) at (rect.north);
        \coordinate (-rectsouth) at (rectprmpt.south);
      \end{scope}
    }
  },
  pics/segmentintermediatesquare/.default={mila-purple-1}{mila-purple-2}{mila-purple}
}

\tikzset{
  pics/segmentintermediatetriangle/.style n args={3}{
    code={
      \begin{scope}[node distance=0pt]
        \begin{scope}[local bounding box=sq]
          \path[draw=none, preaction={fill=delethink-purple!5}, pattern color=black!30]
            (\rectr,0) -- (\sww,0) -- (\sww,\sz) -- (\rectr,\sz)
            arc (90:180:\rectr) -- (0,\rectr) arc (180:270:\rectr) -- cycle;
          \draw[delethink-purple, dashed, line width=1.4\BORDER]
            (\sww,\sz) -- (\rectr,\sz) arc (90:180:\rectr) -- (0,\rectr)
            arc (180:270:\rectr) -- (\rectr,0) -- (\sww,0);
          \path[use as bounding box] (0,0) rectangle (\sww,\sz);
        \end{scope}
        \node[font=\bfseries\large] at (sq.center) {$\mathrm{q}$};

        \node[
          draw=none, minimum height=\sz, minimum width=0.42\rectw,
          inner sep=0pt, outer sep=0pt, right=0pt of sq
        ] (rectprmpt) {};
        \begin{scope}[on background layer]
          \fill[#1] (rectprmpt.south west) rectangle (rectprmpt.north east);
        \end{scope}
        \draw[delethink-purple, line width=1.4\BORDER, dashed]
          (rectprmpt.north west) -- (rectprmpt.north east)
          (rectprmpt.north east) -- (rectprmpt.south east)
          (rectprmpt.south east) -- (rectprmpt.south west);

        \begin{scope}
          \clip (rectprmpt.south west) rectangle (rectprmpt.north east);
        
          \path coordinate (gxW) at ($(rectprmpt.west)!0.19!(rectprmpt.east)$);
          \path coordinate (gxE) at ($(rectprmpt.west)!0.81!(rectprmpt.east)$);
        
          \path coordinate (gy1) at ($(rectprmpt.south)!0.75!(rectprmpt.north)$);
          \fill[white!60] (gxW|-gy1) circle[radius=1.2pt];
          \draw[white!60, line width=\BORDER, line cap=round]
               ($(gxW|-gy1)+(1.6pt,0)$) -- ([xshift=-0pt]gxE|-gy1);
        
          \path coordinate (gy2) at ($(rectprmpt.south)!0.50!(rectprmpt.north)$);
          \fill[white!60] (gxW|-gy2) circle[radius=1.2pt];
          \draw[white!60, line width=\BORDER, line cap=round]
               ($(gxW|-gy2)+(1.6pt,0)$) -- ([xshift=-3pt]gxE|-gy2);
        
          \path coordinate (gy3) at ($(rectprmpt.south)!0.25!(rectprmpt.north)$);
          \fill[white!60] (gxW|-gy3) circle[radius=1.2pt];
          \draw[white!60, line width=\BORDER, line cap=round]
               ($(gxW|-gy3)+(1.6pt,0)$) -- ([xshift=-6pt]gxE|-gy3);
        \end{scope}

        \node[
          draw=#3, line width=1.4\BORDER,
          minimum height=\sz, minimum width=0.5\rectw,
          inner sep=0pt, outer sep=0pt, right=4pt of rectprmpt
        ] (rect) {};
        \begin{scope}[on background layer]
          \fill[#2] (rect.south west) rectangle (rect.north east);
        \end{scope}

        \begin{scope}
          \clip (rect.south west) rectangle (rect.north east);
        
          \def\triH{1.2pt}   %
          \def\triL{1.8
          pt}   %
        
          \coordinate (gxW) at ($(rect.west)!0.18!(rect.east)$);
          \coordinate (gxE) at ($(rect.west)!0.73!(rect.east)$);
        
          \coordinate (gy1) at ($(rect.south)!0.75!(rect.north)$);
          \coordinate (p1)  at ($(gxW|-gy1)$);
          \fill[white]
            ($(p1)+(-\triL,-\triH)$) --
            ($(p1)+(-\triL, \triH)$) --
            ($(p1)+(\triL,0)$) -- cycle;
          \draw[white, line width=\BORDER, line cap=round]
            ($(p1)+(\triL+0.6pt,0)$) -- ([xshift=-3pt]gxE|-gy1);
        
          \coordinate (gy2) at ($(rect.south)!0.50!(rect.north)$);
          \coordinate (p2)  at ($(gxW|-gy2)$);
          \fill[white]
            ($(p2)+(-\triL,-\triH)$) --
            ($(p2)+(-\triL, \triH)$) --
            ($(p2)+(\triL,0)$) -- cycle;
          \draw[white, line width=\BORDER, line cap=round]
            ($(p2)+(\triL+0.6pt,0)$) -- ([xshift=-5pt]gxE|-gy2);
        
          \coordinate (gy3) at ($(rect.south)!0.25!(rect.north)$);
          \coordinate (p3)  at ($(gxW|-gy3)$);
          \fill[white]
            ($(p3)+(-\triL,-\triH)$) --
            ($(p3)+(-\triL, \triH)$) --
            ($(p3)+(\triL,0)$) -- cycle;
          \draw[white, line width=\BORDER, line cap=round]
            ($(p3)+(\triL+0.6pt,0)$) -- (gxE|-gy3);
        \end{scope}

        \coordinate (-sqwest)   at (sq.west);
        \coordinate (-rectwest) at (rect.west);
        \coordinate (-recteast) at (rect.east);
        \coordinate (-east)     at (rect.east);
        \coordinate (-rectnorth) at (rect.north);
        \coordinate (-rectsouth) at (rectprmpt.south);
      \end{scope}
    }
  },
  pics/segmentintermediatetriangle/.default={mila-purple-1}{mila-purple-2}{mila-purple}
}

\tikzset{
  pics/segmentlegend/.style n args={4}{
    code={
      \begin{scope}[node distance=0pt]
        \begin{scope}[local bounding box=sq]
          \path[draw=none, pattern color=black!30]
            (\rectr,0) -- (\sww,0) -- (\sww,0.5\sz) -- (\rectr,0.5\sz)
            arc (90:180:\rectr) -- (0,\rectr) arc (180:270:\rectr) -- cycle;
          \draw[black!70, dashed, line width=0.8\BORDER]
            (\sww,0.5\sz) -- (\rectr,0.5\sz) arc (90:180:\rectr) -- (0,\rectr)
            arc (180:270:\rectr) -- (\rectr,0) -- (\sww,0);
          \path[use as bounding box] (0,0) rectangle (\sww,0.5\sz);
        \end{scope}

        \node[
          draw=none, minimum height=0.5\sz, minimum width=0.04\rectw,
          inner sep=0pt, outer sep=0pt, right=0pt of sq
        ] (rectprmpt) {};
        \draw[black!70, line width=0.8\BORDER, dashed]
          (rectprmpt.north west) -- (rectprmpt.north east)
          (rectprmpt.north east) -- (rectprmpt.south east)
          (rectprmpt.south east) -- (rectprmpt.south west);
        \node[right=1pt of rectprmpt] (txtprompt)
          {\footnotesize \sffamily \textcolor{black!50}{\texttt{Prompt}}};

        \node[
          draw=black!50, line width=1.3\BORDER,
          minimum height=0.5\sz, minimum width=0.25\rectw,
          inner sep=0pt, outer sep=0pt, right=8pt of txtprompt
        ] (rect) {};
        \node[right=1pt of rect] (response)
          {\footnotesize \sffamily \textcolor{black!50}{\texttt{Response}}};

        \node[
          draw=#1, line width=1.3\BORDER, preaction={fill=#1!10},
          minimum height=0.5\sz, minimum width=0.25\rectw,
          inner sep=0pt, outer sep=0pt, right=15pt of response
        ] (query_rect) {};
        \node[right=1pt of query_rect] (query)
          {\footnotesize \sffamily \textcolor{#1}{\texttt{Query}}};

        \node[
          draw=#2, line width=1.3\BORDER, preaction={fill=#2!10},
          minimum height=0.5\sz, minimum width=0.25\rectw,
          inner sep=0pt, outer sep=0pt, right=8pt of query
        ] (reasoning_rect) {};
        \node[right=1pt of reasoning_rect] (reasoning)
          {\footnotesize \sffamily \textcolor{#2}{\texttt{Reasoning}}};

        \node[
          draw=#3, line width=1.3\BORDER, preaction={fill=#3!10},
          minimum height=0.5\sz, minimum width=0.25\rectw,
          inner sep=0pt, outer sep=0pt, right=8pt of reasoning
        ] (summary_rect) {};
        \node[right=1pt of summary_rect] (summary)
          {\footnotesize \sffamily \textcolor{#3}{\texttt{Summary}}};

        \node[
          draw=#4, line width=1.3\BORDER, preaction={fill=#4!10},
          minimum height=0.5\sz, minimum width=0.25\rectw,
          inner sep=0pt, outer sep=0pt, right=8pt of summary
        ] (conclusion_rect) {};
        \node[right=1pt of conclusion_rect] (conclusion)
          {\footnotesize \sffamily \textcolor{#4}{\texttt{Conclusion}}};
      \end{scope}
    }
  },
  pics/segmentlegend/.default={mila-purple-1}{mila-purple-2}{mila-purple}
}

\newcommand{\subfiglongCoT}{
\begin{tikzpicture}
  \path (0,0) pic (seg1) {segmentlongcot={mila-yellow}{delethink-blue}{plot-green}};
  
  \path[fill=black, fill opacity=0, draw opacity=0]
    ([xshift=-0.6cm, yshift=-0.5\sz]seg1-sqwest)
    rectangle ([yshift=0.5\sz]seg1-sqwest);

  \coordinate (mid23) at ($(seg1-east)!0.5!(seg1-sqwest)$);
  \node[anchor=center, fill=none, draw=none, inner xsep=8pt, minimum height=3cm] at (mid23) {};

  \draw[|-|, line width=0.5, draw=black!100, line cap=round]
    ([xshift=1pt, yshift=-25pt]seg1-sqwest) --
    node[midway, below=7pt, align=center, text width=10cm]
      {
       }
    ([xshift=1pt, yshift=-25pt]seg1-east);

  \node[anchor=south west]
    at ([xshift=5pt, yshift=-23pt]current bounding box.north west)
    {\sffamily \large \textbf{Vanilla Reasoning Paradigm}
     \textcolor{black!60}{}};

  \coordinate (NW) at (current bounding box.north west);
  \draw[opacity=0, line width=0pt] (NW) -- ([xshift=16.5cm]NW);
\end{tikzpicture}
}

\newcommand{\subfigdelethink}{
\begin{tikzpicture}
  \path (0,0) pic (seg1) {inftythink_iter_1={mila-yellow}{delethink-blue}{plot-red}};
  
  \path ([yshift=-0.5\sz, xshift=\segap]seg1-east)
    pic (seg2) {inftythink_iter_n-1={mila-yellow}{delethink-blue}{plot-red}};
  \path ([yshift=-0.5\sz, xshift=\segap]seg2-east)
    pic (seg3) {inftythink_iter_n-1={mila-yellow}{delethink-blue}{plot-red}};
  \path ([yshift=-0.5\sz, xshift=\segap]seg3-east)
    pic (seg4) {inftythink_iter_n={mila-yellow}{delethink-blue}{plot-red}{plot-green}};

  \draw[curvedarrow]
    ([yshift=1pt]seg1-north) to[out=40, in=-130, looseness=1.4] ([yshift=-1pt]seg2-rectsouth);
  \draw[curvedarrow]
    ([yshift=1pt]seg2-north) to[out=40, in=-130, looseness=1.4] ([yshift=-1pt]seg3-rectsouth);
  \draw[curvedarrow]
    ([yshift=1pt]seg3-north) to[out=40, in=-130, looseness=1.4] ([yshift=-1pt]seg4-rectsouth);

  \draw[arrowline, shorten >=2pt, shorten <=2pt]
    ([xshift=\segaparrow]seg1-east) -- ([xshift=-0.2\segaparrow]seg2-sqwest);

  \draw[arrowline, shorten >=2pt, shorten <=2pt]
    ([xshift=\segaparrow]seg2-east) -- node[midway, font=\large, fill=white] {\strut$\cdots$}
    ([xshift=-0.2\segaparrow]seg3-sqwest);

  \draw[arrowline, shorten >=2pt, shorten <=2pt]
    ([xshift=\segaparrow]seg3-east) -- ([xshift=-0.2\segaparrow]seg4-sqwest);

  \draw[|-|, line width=0.5, draw=black!100, line cap=round]
    ([xshift=1pt, yshift=-25pt]seg1-sqwest) --
    node[midway, below=7pt, align=center, text width=4cm]
      {\footnotesize \sffamily \textcolor{black!90}{Iter 1\\}}
    ([xshift=1pt, yshift=-25pt]seg1-east);

   \draw[|-|, line width=0.5, draw=black!100, line cap=round]
    ([xshift=1pt, yshift=-25pt]seg2-sqwest) --
    node[midway, below=7pt, align=center, text width=4cm]
      {\footnotesize \sffamily \textcolor{black!90}{Iter 2\\}}
    ([xshift=1pt, yshift=-25pt]seg2-east);

  \draw[|-|, line width=0.5, draw=black!100, line cap=round]
    ([xshift=1pt, yshift=-25pt]seg3-sqwest) --
    node[midway, below=7pt, align=center, text width=4cm]
      {\footnotesize \sffamily \textcolor{black!90}{Iter n-1}}
    ([xshift=1pt, yshift=-25pt]seg3-east);
  
  \draw[|-|, line width=0.5, draw=black!100, line cap=round]
    ([xshift=1pt, yshift=-25pt]seg4-sqwest) --
    node[midway, below=7pt, align=center, text width=4cm]
      {\footnotesize \sffamily \textcolor{black!90}{Iter n}}
    ([xshift=1pt, yshift=-25pt]seg4-east);

  \path ([xshift=-12cm, yshift=-3.3\sz]seg4-sqwest)
    pic (seg5) {segmentlegend={mila-yellow}{delethink-blue}{plot-red}{plot-green}};

  \node[anchor=south west]
    at ([xshift=5pt, yshift=-20pt]current bounding box.north west)
    {\sffamily \large \textbf{InftyThink Reasoning Paradigm}};

  \coordinate (NW) at (current bounding box.north west);
  \draw[opacity=0, line width=0pt] (NW) -- ([xshift=16.5cm]NW);
\end{tikzpicture}
}

\newcommand{\figmethod}{
\begin{figure*}[t]
    \centering
    \begin{minipage}{0.95\textwidth}
      \centering
      \resizebox{\linewidth}{!}{\subfiglongCoT}\par
      \resizebox{\linewidth}{!}{\subfigdelethink}
    \end{minipage}
    \caption{
    \textbf{InftyThink} reasoning paradigm VS. Vanilla reasoning paradigm. \textbf{Upper panel:} The vanilla reasoning paradigm generates a single, continuous long chain-of-thought in one pass. \textbf{Lower panel:} The InftyThink reasoning paradigm decomposes reasoning into multiple iterative rounds, where consecutive iterations are connected via self-generated global summaries.
    }
    \label{fig:inftythink_overview}
\end{figure*}

}

\maketitle

\section{Introduction}
Reinforcement learning has emerged as a powerful paradigm for enhancing the reasoning capabilities of large language models, with code generation serving as a representative task that admits precise, executable verification~\citep{guo2025deepseek, shao2024deepseekmath}. Unlike open-ended text generation, code can be automatically validated against unit tests, providing verifiable rewards that guide policy optimization without human annotation at training time~\citep{NEURIPS2022_8636419d, shojaee2023executionbased}.

However, the effectiveness of this paradigm hinges critically on the quality of the underlying test suites. In practice, obtaining comprehensive unit tests demands substantial human effort, and existing RL-suitable datasets remain limited in both scale and diversity~\citep{chen2021evaluating, austin2021program}. Even in carefully curated benchmarks, each question typically contains only three to five test cases, which cannot reliably distinguish genuinely correct solutions from those that happen to pass by coincidence or handle only common inputs~\citep{liu2023is}. Furthermore, these static golden tests cannot adapt to evolving model capabilities. When tests are overly simple, flawed code may receive undeserved positive rewards; when tests are overly stringent, near-correct solutions are penalized as complete failures~\citep{jeong2025ensuring}. Both scenarios distort the learning signal and limit the potential of RL-based training.

To address the limitations of static rewards, recent work has explored automated test generation~\citep{chencodet, zeng2025acecoder} and self-play frameworks~\citep{wang2025cure, zhao2025absolute} where a single model generates both code and tests. These approaches promise dynamic rewards that adapt to model capabilities, potentially breaking free from static constraints.
Yet they fail to deliver on this promise. Direct generation methods often produce invalid or hallucinated tests~\citep{he2025hardtests}. 
Self-play faces a more fundamental 
\begin{wrapfigure}[25]{r}{0.50\linewidth}
\vspace{-0.3\baselineskip}
\centering
\includegraphics[width=\linewidth]{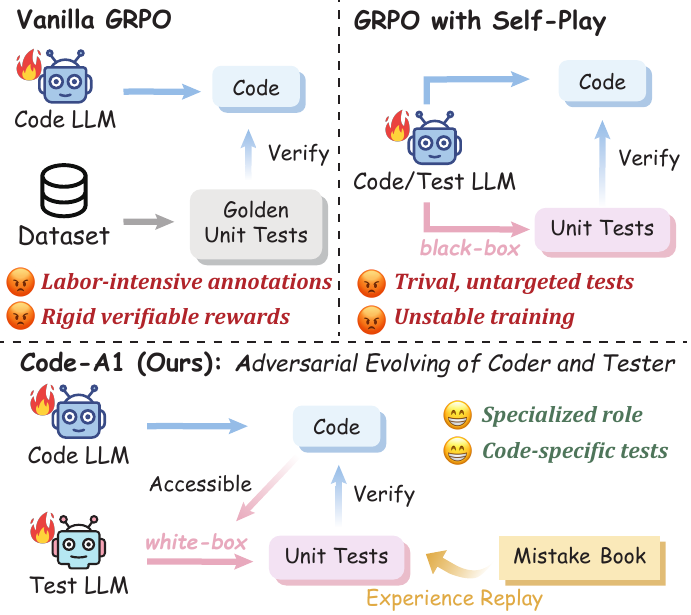}
\captionsetup{font=small, skip=6pt}
\caption{\textbf{Comparison of three training paradigms for code generation.} 
\textit{Vanilla GRPO} relies on static golden tests with rigid rewards. 
\textit{GRPO with Self-Play} unifies code and test generation but must operate in black-box mode to prevent self-collusion. 
\textbf{Code-A1} decouples the two tasks into models with opposing objectives, safely enabling white-box test generation and stable co-evolution via the Mistake Book.}
\label{fig:intro}
\vspace{-0.6\baselineskip}
\end{wrapfigure}
dilemma: when restricted to black-box mode (observing only question descriptions), tests remain generic and miss implementation-specific bugs; when permitted white-box access to candidate code, the model exploits this through self-collusion, generating trivial tests for easy rewards since passing code offsets penalties for weak testing within a unified model~\citep{denison2024sycophancy}. 
To prevent collusion, self-play must restrict test generation to black-box mode, sacrificing the ability to craft targeted tests that probe implementation-specific bugs. This black-box restriction fundamentally undermines dynamic adaptation, as test difficulty becomes decoupled from the actual code being evaluated.

This analysis reveals a key insight: \textit{\textbf{effective verifiable rewards require dynamic interplay between code robustness and test rigor, where test difficulty continuously adapts to challenge the current policy}}.
A good test suite must be valid (executable and correct), sufficiently challenging (exposing real defects), yet not impossibly difficult (providing learnable gradients).
These competing objectives cannot be optimized by a single model or static dataset---they require \emph{adversarial co-evolution} where two specialized agents continuously push each other toward improvement,
with architectural separation that prevents self-collusion while enabling targeted adversarial optimization.

We introduce \textbf{\methodname}, an adversarial reinforcement learning framework that jointly optimizes a Code LLM and a Test LLM with opposing objectives.
Given a question, the Code LLM generates candidate solutions while the Test LLM generates challenging test cases; the two are paired and executed in a sandbox.
The Code LLM receives higher rewards for passing more tests, incentivizing robust and correct solutions.
The Test LLM receives higher rewards for \emph{failing} the code, incentivizing the discovery of edge cases and subtle bugs.
As both models improve, rewards dynamically adapt: stronger code demands harder tests, and harder tests demand stronger code.
This adversarial yet complementary setup enables continuous co-evolution beyond any static performance ceiling.

To stabilize this adversarial dynamics, we introduce several key designs.
First, we decouple the two tasks into separate models, eliminating the self-collusion risks inherent in self-play and enabling safe white-box adversarial optimization.
Second, we design a composite reward for the Test LLM that balances \emph{validity} (tests must execute correctly) with \emph{adversarial difficulty} (tests should expose defects), avoiding both trivial and impossible tests.
Third, we maintain a \emph{Mistake Book}---an experience replay buffer that records historically failed tests for each question---ensuring that resolved bugs are not forgotten and providing stable baselines for reward computation.

We conduct extensive experiments on Qwen2.5-Coder models (1.5B, 3B, 7B).
On code generation benchmarks, \methodname consistently outperforms both models trained on human-annotated golden tests and self-play baselines across all scales. 
On test generation, the results reveal remarkable efficiency: the 3B model achieves a Mul score of 15.29, surpassing the 7B base model (14.72), demonstrating that adversarial co-evolution discovers bug-revealing patterns more effectively than parameter scaling alone.

Our contributions can be summarized as follows:

\begin{itemize}
\item We introduce adversarial co-evolution into code RL, enabling dynamic and adaptive verifiable rewards that eliminate reliance on static human-annotated test suites.
\item We develop \methodname, comprising dual-policy optimization with opposing objectives, validity-aware reward shaping for the Test LLM, and a Mistake Book mechanism for stable experience replay.
\item We demonstrate empirically that \methodname matches or exceeds the performance of RL with static golden tests on code generation benchmarks, while simultaneously producing a Test LLM capable of generating high-quality, bug-revealing tests.
\end{itemize}

\section{Related Works}
\subsection{Reinforcement Learning for Code Reasoning}

Reinforcement Learning (RL) effectively bridges the gap left by traditional supervised learning, which primarily focuses on similarity at the token level, by directly optimizing for the functional correctness of code. Unlike general text generation, code generation inherently possesses an executable verification environment, enabling the utilization of compiler feedback and unit test outcomes as reward signals. This characteristic naturally aligns code generation with reinforcement learning paradigms. Early explorations utilized architectures comprising Actor and Critic networks, guiding models to generate compliant code by incorporating dense feedback signals derived from test case pass rates \citep{NEURIPS2022_8636419d}. Building on this, methods based on Proximal Policy Optimization (PPO) have been widely applied to directly integrate discrete execution feedback~\citep{zhang2024o1codero1replicationcoding}. These approaches translate compilation rates and test pass rates into reward values, while employing KL divergence constraints to stabilize the training process \citep{shojaee2023executionbased}.

To address the issue of sparse reward signals in code tasks—where programs are typically binary (correct or incorrect) and lack intermediate states—researchers have introduced optimization strategies based on granular feedback. The RLTF framework leverages error signals across various levels (e.g., compilation, runtime, and logic errors) to provide immediate feedback at multiple levels of detail, thereby guiding model exploration more precisely during online training \citep{liu2023rltf}. Furthermore, given the high computational overhead and potential instability associated with online RL algorithms like PPO, efficient alignment methods based on ranking have also been adapted for code tasks. These approaches eliminate the need for an explicit reward model; instead, they rank and select sampled candidate code based on test outcomes, achieving effective alignment with execution feedback while significantly reducing training costs \citep{shen2023pangu}.

\subsection{Unit tests Generation for Code Reasoning}

Unit test synthesis is increasingly automated by Large Language Models (LLMs) to overcome the high cost and scalability issues of manual creation~\citep{yang2025swesmith, tip2025llmorpheus}. A common approach involves generating test cases from a question description and then using a proxy solution from a more capable model to filter out hallucinations and ensure quality~\citep{zeng2025acecoder, jain2025testgeneval}. More advanced strategies focus on generating difficult test cases, such as ``hacking inputs'' designed to induce timeouts, by prompting models to write ``test generator programs'' and using oracle programs to validate the outputs~\citep{he2025hardtests, altmayer2025coverup, hossain2025togll}. Furthermore, co-evolutionary frameworks enable a model to act as both code and test generator, engaging in self-play for mutual, unsupervised improvement~\citep{wang2025cure, zhao2025absolute, chen2025multi, lu2025searchselfplay}.

These automatically synthesized tests serve two primary purposes. First, they provide a reliable and scalable reward signal for reinforcement learning, where a code's pass rate on these tests directly fine-tunes the model~\citep{zhang2024o1}. Second, at inference time, they enable self-verification in agentic workflows. For instance, in a ``Best-of-N (BoN)'' strategy, the candidate solution that passes the most self-generated tests is selected, which also underpins more complex iterative debugging and refinement processes~\citep{wang2025cure}.

\begin{figure*}[h]
    \centering
    \includegraphics[width=0.95\textwidth]{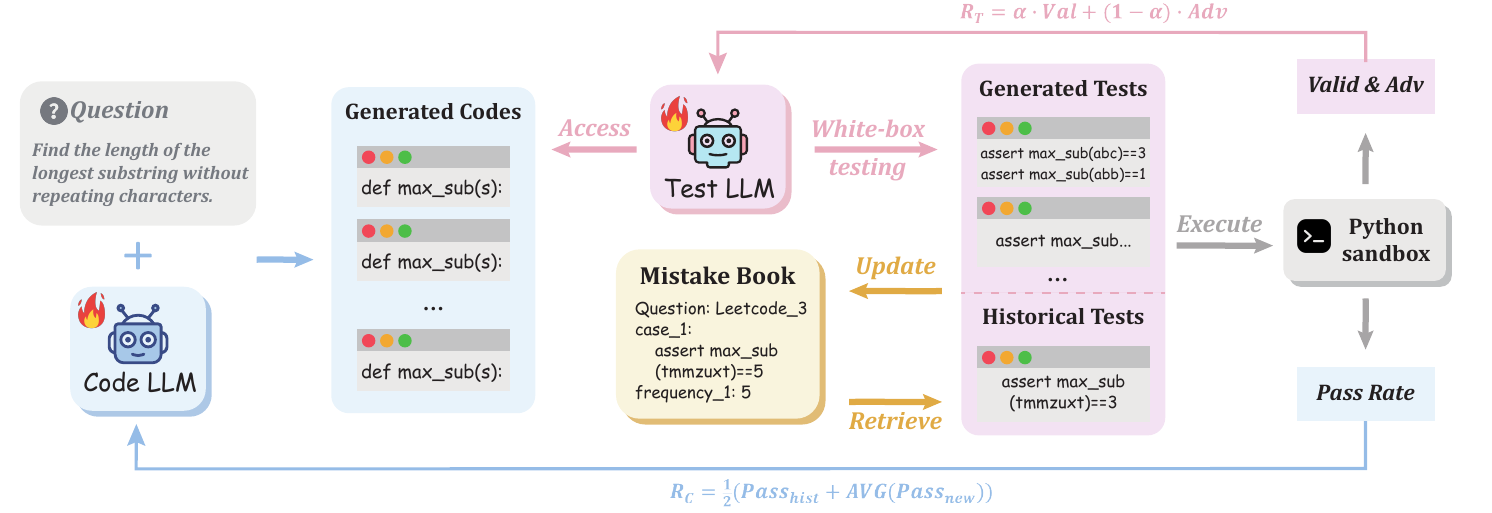}
    \caption{
    \textbf{Overview of the \methodname training framework.} The Code LLM generates solutions accessible to the Test LLM for white-box testing. Generated tests are validated and merged with historical tests from the Mistake Book. The Code LLM is rewarded for passing more tests; the Test LLM is rewarded for exposing more defects, enabling curriculum-aware adversarial co-evolution.
    }
    \label{fig:method}
\end{figure*}

\section{Methods}
\label{sec:methods}

In this section, we present the \methodname framework for adversarial co-evolution of code and test generation. We begin with problem formulation and output format constraints (Section~\ref{sec:formulation}). We then describe the adversarial rollout procedure (Section~\ref{sec:rollout}) and the Mistake Book mechanism for experience replay (Section~\ref{sec:mistake_book}). Finally, we detail the reward design (Section~\ref{sec:reward}) and policy optimization (Section~\ref{sec:optimization}).

\subsection{Problem Formulation}
\label{sec:formulation}

We consider function-level code generation, where a model receives a question description $Q$ containing a function signature and natural language specification, and outputs a complete function body. Let $\pi_C$ denote the Code LLM and $\pi_T$ denote the Test LLM. Given a dataset $\mathcal{D} = \{(Q_i, C_i)\}_{i=1}^{|\mathcal{D}|}$ where $C_i$ is the ground-truth solution, our goal is to jointly optimize both models through adversarial interaction.

The Code LLM generates candidate solutions $\hat{C} \sim \pi_C(\cdot \mid Q)$ that should be syntactically correct and satisfy the specification. The Test LLM generates a set of $K$ test cases $\hat{T} = \{\hat{t}_1, \ldots, \hat{t}_K\} \sim \pi_T(\cdot \mid Q, \hat{C})$ conditioned on both the question and a candidate solution. Each test case follows the assertion format \texttt{assert func(*args) == answer}, where \texttt{func} is the target function, \texttt{*args} are input arguments, and \texttt{answer} is the predicted output. This structured format enables reliable extraction via abstract syntax tree parsing, reducing reward noise from formatting errors. We require exactly $K$ test cases per response to prevent trivial convergence to single-test outputs and discourage brute-force generation for reward hacking.

\subsection{Adversarial Rollout}
\label{sec:rollout}

At each training step, we sample a batch of questions from $\mathcal{D}$ and perform adversarial rollout.

\paragraph{Step 1: Code Generation.} For each question $Q$, the Code LLM generates $M$ candidate solutions $\{\hat{C}_1, \ldots, \hat{C}_M\}$ via sampling.

\paragraph{Step 2: Test Generation.} For each candidate solution $\hat{C}_m$, the Test LLM generates $N$ test suites $\{\hat{T}_{m,1}, \ldots, \hat{T}_{m,N}\}$, conditioned on both $Q$ and $\hat{C}_m$. Conditioning on the candidate solution enables the Test LLM to craft targeted tests that probe potential weaknesses.

\paragraph{Step 3: Test Validation.} We extract function calls from each generated test and execute them against the ground-truth solution $C_{GT}$. A test is deemed \emph{valid} if: (i) the function call executes without error, (ii) the call is unique within the test suite, and (iii) the the predicted answer is correct. For tests with incorrect predicted answers, we replace the prediction with the ground-truth return value, retaining the test to enrich coverage. Other Invalid tests are discarded.

\paragraph{Step 4: Code Evaluation.} We concatenate each solution $\hat{C}_m$ with the validated test suites and execute in a sandboxed environment. The pass rate serves as the basis for reward computation.

\subsection{Mistake Book}
\label{sec:mistake_book}

A key challenge in adversarial training is instability: a weak Test LLM may generate trivial tests, providing inflated rewards that mislead the Code LLM. Conversely, a strong Test LLM may generate tests so difficult that learning signals vanish. To stabilize training and track capability evolution, we introduce the \emph{Mistake Book}, a per-question experience replay buffer~\citep{zhan2025exgrpo}.

\paragraph{Structure.} Mistake Book $\mathcal{B}$ maintains a mapping from each question $Q_i$ to a set of historically failed tests:
\begin{equation}
\mathcal{B}: Q_i \mapsto T_i^{\text{hist}} = \{t \mid \hat{C} \text{ failed } t \text{ in previous iters}\}
\end{equation}

\paragraph{Update Rule.} After each training step, we update $\mathcal{B}$ dynamically: newly generated tests that the current candidate solutions fail (NewFails) are added to $T_i^{\text{fail}}$, while historical tests that the solutions now pass (NewPasses) are removed. This ensures that $\mathcal{B}$ reflects the frontier of model capability, containing exactly those tests that remain challenging given current Code LLM performance.

\paragraph{Role in Training.} The Mistake Book serves three purposes. First, historical tests provide a stable baseline for reward computation, reducing variance caused by stochastic test generation. Second, the gap between historical and new test pass rates provides a curriculum signal that reveals whether the Test LLM is generating progressively harder tests. Third, re-evaluating against historical failures prevents forgetting, ensuring that previously fixed bugs are not reintroduced as training proceeds.

\subsection{Reward Design}
\label{sec:reward}

We assign rewards at the trajectory level, with opposing objectives for the two models.

\paragraph{Code LLM Reward.} The Code LLM should produce solutions that are both correct and robust. We evaluate each candidate solution against two test sources: historical failures from the Mistake Book and newly generated tests from the Test LLM. Let $\text{Pass}_{\text{hist}}(\hat{C}_m)$ denote the pass rate on historical tests $T^{\text{hist}}$, and $\textsc{Avg}(\text{Pass}_{\text{new}}(\hat{C}_m))$ denote the average pass rate on newly generated test suites. The reward is:
\begin{equation}
R_C(\hat{C}_m) = 
\begin{cases}
\frac{1}{2}\left(\text{Pass}_{\text{hist}} + \textsc{Avg}\left(\text{Pass}_{\text{new}}\right)\right) & \text{if } T^{\text{hist}} \neq \emptyset \\[6pt]
\text{Pass}_{\text{new}} & \text{otherwise}
\end{cases}
\end{equation}
When historical failures exist, averaging the two pass rates ensures that the Code LLM cannot achieve high rewards by merely passing new tests while regressing on previously challenging cases.

\paragraph{Test LLM Reward.} The Test LLM faces a fundamental tension: tests must be \emph{valid} (syntactically executable, correct and unique) yet \emph{adversarial} (capable of exposing defects). We design a composite reward to balance these objectives. The validity reward $R_{T}^{\text{val}}(\hat{T}) = \text{Valid}(\hat{T})$ measures the fraction of generated tests that pass validation, implicitly encouraging format compliance. The adversarial reward measures whether new tests are harder than historical ones:
\begin{equation}
R_{T}^{\text{adv}}(\hat{T}) = 
\begin{cases}
\frac{1}{2}\left(\text{Pass}_{\text{hist}} - \text{Pass}_{\text{new}} + 1\right) & \text{if } T^{\text{hist}} \neq \emptyset \\[6pt]
1 - \text{Pass}_{\text{new}} & \text{otherwise}
\end{cases}
\end{equation}
When $\text{Pass}_{\text{hist}} > \text{Pass}_{\text{new}}$, the new tests expose defects that historical tests missed, yielding higher reward. When $\text{Pass}_{\text{hist}} < \text{Pass}_{\text{new}}$, the new tests are easier than historical ones, incurring penalty. The final reward balances validity and adversarial objectives:
\begin{equation}
R_T(\hat{T}) = \alpha \cdot R_{T}^{\text{val}}(\hat{T}) + (1 - \alpha) \cdot R_{T}^{\text{adv}}(\hat{T})
\label{eq:4}
\end{equation}
where $\alpha \in [0, 1]$ controls the trade-off. Setting $\alpha$ too high encourages trivial but valid tests; setting $\alpha$ too low risks invalid adversarial tests. We study this trade-off in ablations.

\subsection{Policy Optimization}
\label{sec:optimization}

We adopt Group Relative Policy Optimization (GRPO)~\citep{shao2024deepseekmath} with token-level loss aggregation~\citep{yu2025dapo} for both models. For a question $Q$ with $G$ sampled trajectories, the GRPO objective is:
\begin{equation}
\mathcal{J}(\theta) = \mathbb{E}\left[\frac{1}{G}\sum_{g=1}^{G} \hat{A}_g \sum_{t=1}^{|o_g|} \log \pi_\theta(o_g^t \mid o_g^{<t})\right]
\end{equation}
where $\hat{A}_g = (R_g - \mu) / \sigma$ is the normalized advantage computed from group statistics.

\noindent
\begin{minipage}[t]{0.48\linewidth}
\vspace{0pt}

\paragraph{Asymmetric Sampling.} The Code LLM generates $M$ solutions per question, while the Test LLM generates $M \times N$ test suites (N suites per solution). To balance training compute between models, we select only the top-$\ell$ test suite groups with highest reward variance for the Test LLM update (see details in Appendix~\ref{appendix:topvar}), setting $\ell \times N = M$. This prioritizes high-learning-value samples while maintaining synchronized training steps.

\paragraph{Training Dynamics.} The adversarial setup creates a natural curriculum. In early training, both models have limited capability, and the Test LLM generates simple tests that provide achievable targets for the Code LLM without overwhelming gradients. As training progresses, the Code LLM improves and passes most historical tests, forcing the Test LLM to generate harder tests to earn rewards, which in turn raises the bar for the Code LLM. Eventually, the two models reach an equilibrium where further improvement requires genuine capability gains rather than exploitation of weak opponents. By decoupling code and test generation into separate models with opposing objectives, \methodname avoids the reward hacking risks inherent in white-box self-play while enabling continuous co-evolution.

\end{minipage}\hfill
\begin{minipage}[t]{0.50\linewidth}
\vspace{0pt}

\refstepcounter{algorithm}
\hrule
\vspace{3pt}
\small
\textbf{Algorithm \thealgorithm} \ \methodname
\label{alg:code-a1}
\vspace{3pt}
\hrule
\vspace{3pt}
\begin{algorithmic}[1]
\REQUIRE Coder $\pi_C$, Tester $\pi_T$, dataset $\mathcal{D}$, Mistake Book $\mathcal{B}$
\STATE Initialize Mistake Book $\mathcal{B} \leftarrow \emptyset$
\WHILE{not converged}
    \FOR{$(Q, C_{GT}) \in \mathcal{D}$}
        \STATE \textcolor{Green}{\textit{// Rollout}}
        \STATE $\{\hat{C}_m\}_{m=1}^{M} \sim \pi_C(\cdot \mid Q)$
        \STATE \textcolor{Green}{\textit{// Retrieve Historical Failures}}
        \STATE $T^{\text{hist}} \leftarrow \mathcal{B}(Q)$
        \STATE \textcolor{Green}{\textit{// Test Generation \& Validation}}
        \FOR{$m = 1$ to $M$}
            \STATE $\{\hat{T}_{m,n}\}_{n=1}^{N} \sim \pi_T(\cdot \mid Q, \hat{C}_m)$
            \STATE $T_{m,n} \leftarrow \textsc{Validate}(\hat{T}_{m,n}, C_{GT})$
        \ENDFOR
        \STATE $\text{Pass}^{(m)} \leftarrow \textsc{Execute}(\hat{C}_m, \{T_{m,n}\}_{n=1}^{N} \cup T^{\text{hist}})$
        \STATE \textcolor{Green}{\textit{// Reward Computation}}
        \STATE $R_C^{(m)} \leftarrow f(\text{Pass}_{\text{hist}}^{(m)}, \text{Pass}_{\text{new}}^{(m)})$
        \STATE $R_T^{(m,n)} \leftarrow \alpha \cdot \text{Valid} + (1-\alpha) \cdot \text{Adv}$
        \STATE \textcolor{Green}{\textit{// Update Mistake Book}}
        \STATE $\mathcal{B}(Q) \leftarrow \mathcal{B}(Q) \cup \text{NewFails} \setminus \text{NewPasses}$
        \STATE \textcolor{Green}{\textit{// Policy Update via GRPO}}
        \STATE Update $\pi_C$ with $\{R_C^{(m)}\}$
        \STATE Update $\pi_T$ with $\textsc{TopVar}(\{R_T^{(m,n)}\})$
    \ENDFOR
\ENDWHILE
\end{algorithmic}
\vspace{3pt}
\hrule
\end{minipage}

\section{Experiments}
\label{sec:experiments}
\subsection{Experimental Setup}
\label{sec:setup}

\begin{table}[h]
  \centering
  \small
  \caption{\textbf{Performance comparison of Code LLMs.}
  The best results are highlighted in bold, and the second best results are underlined.}
  \label{tab:code}
  \setlength{\tabcolsep}{10pt}
  \begin{tabular}{llcccc}
    \toprule
    \multicolumn{1}{l}{\multirow{2}{*}{\textbf{Code LLM}}} &
    \multirow{2}{*}{\textbf{Method}} &
    \textbf{HumanEval$^+$} &
    \textbf{MBPP$^+$} &
    \textbf{BigCodeBench} &
    \multirow{2}{*}{\textbf{Avg}} \\
    & & avg@32 & avg@32 & avg@32 & \\
    \midrule
    Qwen2.5-Coder-1.5B-Instruct & / & 63.42 & 60.87 & 29.34 & 51.21 \\
    & Golden Tests & \underline{71.15} & 63.30 & \underline{34.23} & \underline{56.23} \\
    & Self-Play & 70.64 & \textbf{63.54} & 33.47 & 55.88 \\
    \rowcolor{gray!20} & \methodname & \textbf{72.69} & \underline{63.33} & \textbf{34.82} & \textbf{56.95} \\
    \midrule
    Qwen2.5-Coder-3B-Instruct & / & 77.63 & 63.12 & 41.78 & 60.84 \\
    & Golden Tests & \underline{81.96} & \underline{68.05} & \underline{45.41} & \underline{65.14} \\
    & Self-Play & 81.86 & 67.06 & 45.09 & 64.67 \\
    \rowcolor{gray!20} & \methodname & \textbf{83.52} & \textbf{69.07} & \textbf{45.85} & \textbf{66.15} \\
    \midrule
    Qwen2.5-Coder-7B-Instruct & / & 83.69 & 71.95 & 49.41 & 68.35 \\
    & Golden Tests & 84.68 & 74.16 & \underline{52.28} & 70.37 \\
    & Self-Play & \underline{84.70} & \underline{74.23} & 52.25 & \underline{70.39} \\
    \rowcolor{gray!20} & \methodname & \textbf{85.21} & \textbf{74.50} & \textbf{52.46} & \textbf{70.72} \\
    \bottomrule
  \end{tabular}
\end{table}

\begin{table*}[h]
  \centering
  \small
  \caption{\textbf{Performance comparison of Test LLMs.} The best results are highlighted in bold, and the second best results are underlined.
  }
  \label{tab:test}%
  \setlength{\tabcolsep}{6pt}
    \begin{tabular}{llccccccc}
    \toprule
    \multicolumn{1}{l}{\multirow{2}[2]{*}{\textbf{Test LLM}}} & \multirow{2}[2]{*}{\textbf{Method}} & \multicolumn{6}{c}{\textbf{UnLeakedTestBench}} & \multirow{2}[2]{*}{\textbf{Mul}} \\
    \cmidrule(lr){3-8}
          &       & \multicolumn{1}{c}{pass@1} & \multicolumn{1}{c}{pass@2} & \multicolumn{1}{c}{pass@5} & \multicolumn{1}{c}{mut@1} & \multicolumn{1}{c}{mut@2} & \multicolumn{1}{c}{mut@5} &  \\
    \midrule
    \multicolumn{1}{l}{Qwen2.5-Coder-1.5B-Instruct} & /     & 18.34 & 16.91 & 16.29 & 17.10 & 20.35 & 22.30 & 3.63 \\
          & SFT   & 16.01 & 15.16 & 14.76 & 14.53 & 27.57 & 29.45 & 4.35 \\
          & Self-Play & 24.12 & 23.37 & 23.39 & 14.82 & 26.98 & 28.91 & \underline{6.76} \\
          \rowcolor{gray!20} & \methodname & 27.42 & 25.68 & 27.05 & 17.31 & 23.57 & 26.41 & \textbf{7.14} \\
    \midrule
    \multicolumn{1}{l}{Qwen2.5-Coder-3B-Instruct} & /     & 19.77 & 20.50 & 20.93 & 28.35 & 29.99 & 42.55 & 8.91 \\
          & SFT   & 25.40 & 25.71 & 23.51 & 26.77 & 29.08 & 36.29 & 8.53 \\
          & Self-Play & 25.84 & 29.65 & 29.64 & 29.66 & 38.88 & 50.92 & \underline{15.09} \\
          \rowcolor{gray!20} & \methodname & 27.83 & 30.31 & 30.86 & 26.44 & 36.96 & 49.56 & \textbf{15.29} \\
    \midrule
    \multicolumn{1}{l}{Qwen2.5-Coder-7B-Instruct} & /     & 27.71 & 28.64 & 28.73 & 30.36 & 36.07 & 51.25 & 14.72 \\
          & SFT   & 29.32 & 29.94 & 28.72 & 31.33 & 41.01 & 50.85 & 14.60 \\
          & Self-Play & 32.98 & 34.74 & 35.13 & 31.42 & 38.54 & 55.57 & \underline{19.52} \\
          \rowcolor{gray!20} & \methodname & 36.79 & 37.87 & 37.15 & 35.50 & 36.60 & 53.14 & \textbf{19.74} \\
    \bottomrule
    \end{tabular}%
\end{table*}%

\paragraph{Implementation.} We use Qwen2.5-Coder-Instruct models~\citep{hui2024qwen2} at three scales (1.5B, 3B, 7B) as base models. The Code LLM and Test LLM are initialized from the same checkpoint and trained jointly on 9,688 hard-difficulty questions from KodCode-V1~\citep{xu2025kodcode}. For the Test LLM, we apply supervised fine-tuning before RL to establish the assertion format. During rollout, both models sample 8 responses per question with temperature 1.0. The Test LLM generates $K=5$ test cases per response. We set $\ell=1$ for the Test LLM to balance training compute and $\alpha=0.5$ for the validity-adversarial trade-off. Training runs for 111 steps with GRPO. Additional implementation details including prompt design, sandbox configuration, and Mistake Book structure are provided in Appendix~\ref{appendix:experimental_details}.

\paragraph{Baselines.} We compare \methodname against two groups of baselines. For code generation, we consider:  \textbf{Base}, the original Qwen2.5-Coder-Instruct without RL; \textbf{Golden Tests}, the Code LLM trained via GRPO using human-annotated tests as verifiable rewards; and  \textbf{Self-Play}, which employs a single model for both code and test generation, with input isolation restricting test generation to question descriptions only to prevent reward hacking. For test generation, we additionally include  \textbf{SFT}, which trains the Test LLM with supervised fine-tuning only. Implementation details for SFT and Self-Play are provided in Appendix~\ref{appendix:sft} and~\ref{appendix:selfplay}.

\paragraph{Evaluation.} We evaluate code generation on HumanEval$^+$\citep{liu2023is}, MBPP$^+$~\citep{liu2023is}, and BigCodeBench~\citep{zhuo2025bigcodebench}, and test generation on a 10\% subset of UnLeakedTestBench~\citep{huang2025benchmarking}. We sample 32 responses for Code LLM and 5 for Test LLM, with temperature 0.7 and top-$p$ 0.95. We report avg@32 for code generation, and pass@$k$ (test accuracy) and mut@$k$ (mutation score) for test generation. To assess comprehensive performance, we additionally introduce Avg (the mean of code generation scores) and Mul ($pass@k \times mut@k$), a composite metric balancing test validity and adversarial power (Appendix~\ref{appendix:metrics}).

\subsection{Main Results}
\label{sec:main_results}

\paragraph{Code Generation.} Table~\ref{tab:code} presents results across three model scales.
\methodname consistently achieves the highest average scores, outperforming both the Golden Tests baseline trained on human annotations and the Self-Play approach.
The advantage is most pronounced at smaller scales: on the 1.5B model, \methodname achieves 56.95\% average accuracy compared to 56.23\% for Golden Tests and 55.88\% for Self-Play.
This gap stems from the fundamental difference in testing paradigms. Self-Play must operate in black-box mode to prevent self-collusion, generating tests solely from question descriptions. In contrast, \methodname's decoupled architecture safely enables white-box testing, where the Test LLM inspects candidate code to craft targeted adversarial tests. This produces richer, on-policy reward signals that drive the Code LLM to develop robustness against precise vulnerabilities rather than generic edge cases.
We further validate this advantage against CURE~\citep{wang2025cure} in Appendix~\ref{appendix:cure}, where \methodname-3B surpasses their 7B model.

\paragraph{Test Generation.} Table~\ref{tab:test} reveals that \methodname significantly outperforms both Base and SFT models on test generation, with Test LLM evolving from simple validity to high discriminatory power. 
The results demonstrate a striking efficiency gain: the 3B model trained with \methodname achieves a Mul score of 15.29, surpassing the 7B Base model (14.72). 
This indicates that adversarial co-evolution discovers bug-revealing patterns more effectively than parameter scaling alone. 
Unlike SFT which produces valid but generic tests, \methodname drives the Test LLM to synthesize targeted tests that expose implementation-specific flaws. 
The consistent improvement across all scales (1.5B: 7.14 vs 4.35, 3B: 15.29 vs 8.53, 7B: 19.74 vs 14.60 in Mul) confirms that dynamic adversarial pressure is essential for developing discriminatory test generation capability.

\subsection{Ablation Studies}
\label{sec:ablation}

\begin{figure*}[h]
    \centering
    \includegraphics[width=0.48\textwidth]{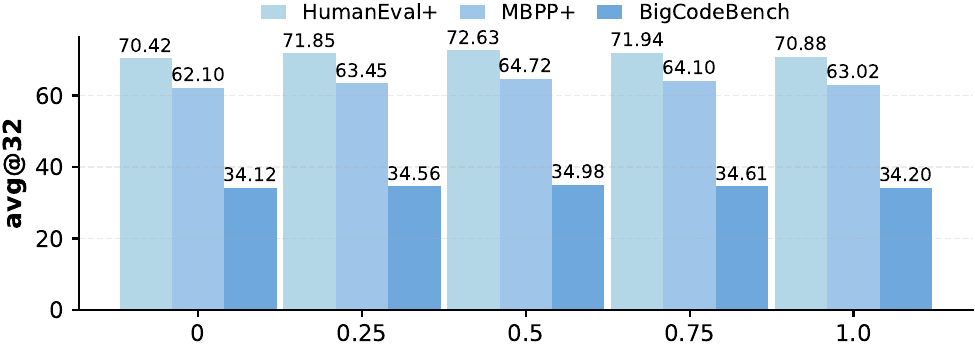}
    \hfill
    \includegraphics[width=0.48\textwidth]{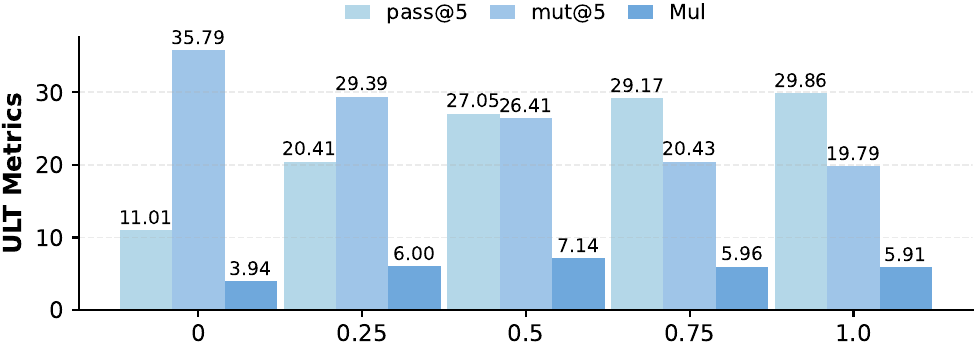}
    \caption{Impact of trade-off weight $\alpha$ on model capabilities using Qwen2.5-Coder-1.5B-Instruct. Left: Code generation performance (avg@32) on three benchmarks. Right: Test generation metrics on UnLeakedTestBench.}
    \label{fig:alpha}
\end{figure*}

\paragraph{Trade-off Weight $\alpha$.} The weight $\alpha$ in Eq.~\ref{eq:4} balances validity and adversarial difficulty in the Test LLM reward. As shown in Figure~\ref{fig:alpha}, $\alpha=0.5$ achieves optimal performance on both Code LLM (56.95 Avg) and Test LLM (7.14 Mul) metrics. When $\alpha$ is too low (e.g., $\alpha=0$), the reward overemphasizes difficulty while neglecting correctness. Although our mechanism corrects erroneous predictions using ground truth, the lack of intrinsic correctness incentives causes policy entropy to spike; the Test LLM generates challenging tests by chance rather than learning robust patterns, resulting in unstable training and poor Mul (3.94). Conversely, when $\alpha$ is too high (e.g., $\alpha=1.0$), Test LLM converges to trivial tests that provide insufficient learning gradients for Code LLM. The balanced setting $\alpha=0.5$ maintains stability while preserving adversarial pressure.

\paragraph{Multi-Objective Optimization.} We investigate whether removing answer prediction simplifies optimization. In the ``w/o predicted answer'' setting (Table~\ref{tab:ablation}), the Test LLM generates only function calls, relying on oracles for expected outputs. While this accelerates convergence toward difficult tests, it fails to significantly improve Code LLM performance and severely degrades test generation capability (pass@5 drops to 11.18\%, below the Base model). This confirms that requiring the Test LLM to predict answers serves dual purposes: preventing impossible tests that eliminate learning gradients, and maintaining the Test LLM's intrinsic capability to synthesize complete unit tests.

\begin{table*}[h]
  \centering
  \small
  \caption{
  \textbf{Ablation study: } removing referable code, predicted answer, or Mistake Book each degrades performance, confirming the contribution of white-box access, multi-objective optimization, and experience replay.
  }
  \setlength{\tabcolsep}{9.5pt}
    \resizebox{\columnwidth}{!}{
    \begin{tabular}{lccccccc}
    \toprule
    \multirow{2}[2]{*}{\textbf{Settings}} & \multicolumn{1}{c}{\textbf{HumanEval$^+$}} & \multicolumn{1}{c}{\textbf{MBPP$^+$}} & \multicolumn{1}{c}{\textbf{BigCodeBench}} & \multirow{2}[2]{*}{\textbf{Avg}} & \multicolumn{2}{c}{\textbf{UnLeakedTestBench}} & \multirow{2}[2]{*}{\textbf{Mul}} \\
    \cmidrule(lr){6-7}
          & \multicolumn{1}{c}{avg@32} & \multicolumn{1}{c}{avg@32} & \multicolumn{1}{c}{avg@32} &       & \multicolumn{1}{c}{pass@5} & \multicolumn{1}{c}{mut@5} &  \\
    \midrule
    \rowcolor{gray!20}{\methodname} & \textbf{72.69} & 63.33 & \textbf{34.82} & \textbf{56.95} & 27.05 & 26.41 & \textbf{7.14} \\
        {- w/o predicted answer} & \underline{70.69} & \textbf{64.47} & \underline{34.60} & \underline{56.59} & 11.18 & 21.72 & 2.43 \\
        {- w/o referable code} & 68.66 & 62.99 & 33.34 & 55.00 & 23.77 & 23.52 & 5.59 \\
        {- w/o Mistake Book} & 69.66 & \underline{63.94} & 33.73 & 55.78 & 25.50 & 27.41 & \underline{6.99} \\
    \bottomrule
    \end{tabular}%
    }
  \label{tab:ablation}%
\end{table*}%

\paragraph{Conditioning on Candidate Code.} Removing candidate code from the Test LLM input (``w/o referable code'' setting in Table~\ref{tab:ablation}) causes HumanEval$^+$ to drop from 72.69\% to 68.66\%. This ablation directly demonstrates the value of white-box access. Black-box generation yields broad tests that miss implementation-specific bugs. In \methodname, the decoupled architecture transforms white-box access from a collusion risk into a debugging advantage: the Test LLM inspects code structure to craft on-policy tests targeting specific logical weaknesses, providing sharper curriculum-like signals that black-box methods cannot replicate.

\begin{figure*}[h]
\centering
    \begin{subfigure}[t]{0.45\textwidth}
        \centering
        \includegraphics[width=\textwidth]{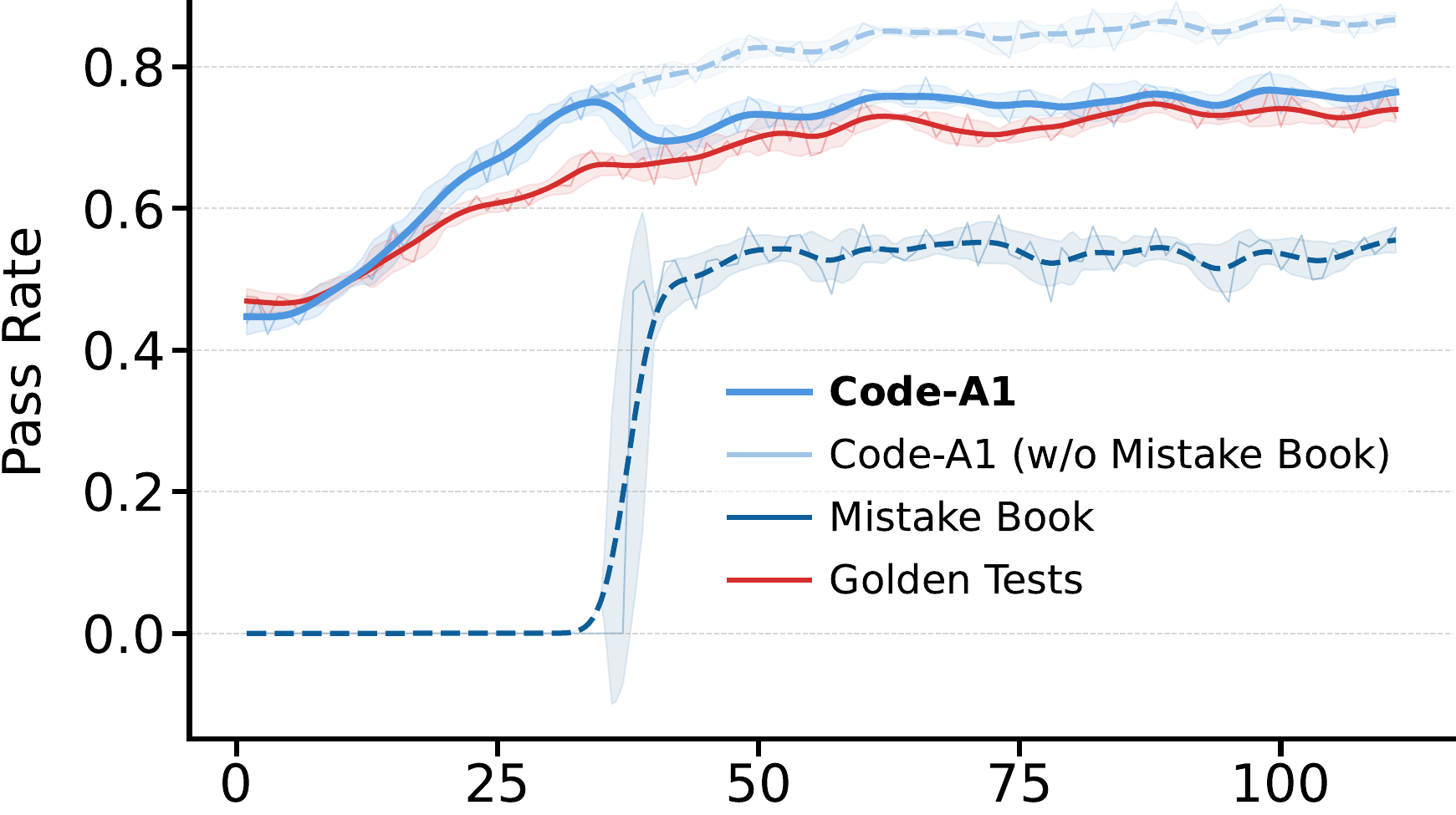}
        \caption{Pass rate over training steps.}
        \label{fig:mistake:a}
    \end{subfigure}
    \hspace{0.02\textwidth}
    \begin{subfigure}[t]{0.45\textwidth}
        \centering
        \includegraphics[width=\textwidth]{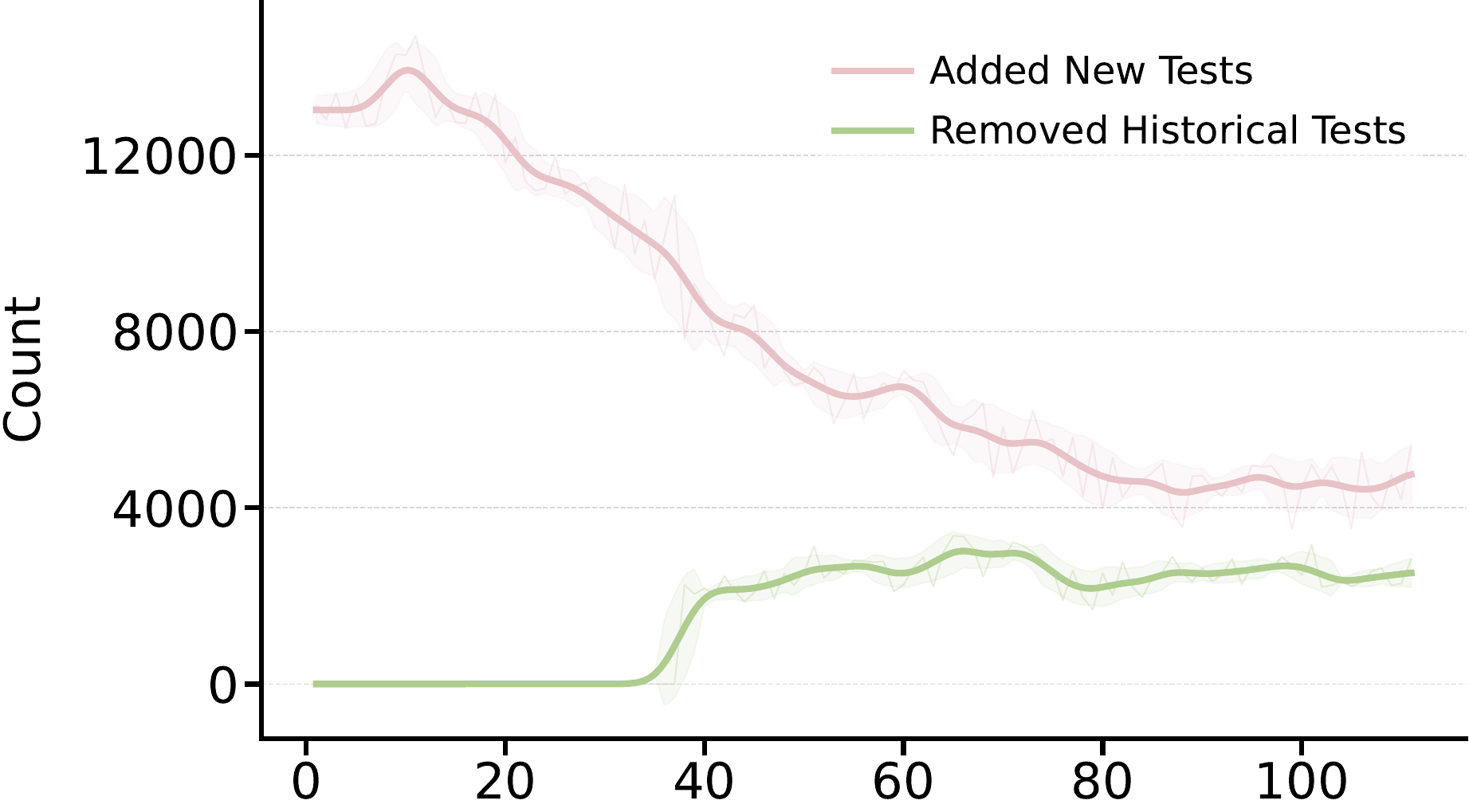}
        \caption{Dynamics of Mistake Book over training steps.}
        \label{fig:mistake:b}
    \end{subfigure}
    \caption{\textbf{Effect of Mistake Book.} (a) Pass rate on newly generated tests, historical tests, and Golden Tests baseline over training. (b) Number of tests added to and removed from the Mistake Book per step.}
    \label{fig:mistake}
\end{figure*}

\paragraph{Mistake Book.} Ablating the Mistake Book (``w/o Mistake Book'' setting in Table~\ref{tab:ablation}) degrades Code LLM performance (HumanEval$^+$ drops from 72.69\% to 69.66\%) while minimally affecting Test LLM (Mul: 7.14 → 6.99). This asymmetry reveals the mechanism's primary role: preventing catastrophic forgetting in the Code LLM. Without re-evaluation against historical failures, the model regresses on previously resolved bugs while overfitting to current adversarial attacks.
Figure~\ref{fig:mistake} further visualizes how the Mistake Book stabilizes adversarial co-evolution. In Figure~\ref{fig:mistake:a}, the pass rate on historical tests (dark blue) begins tracking at Step 38 after the first epoch populates the buffer. While enforcing historical correctness initially lowers the composite pass rate compared to the ablation without Mistake Book (solid vs. dashed blue), this prevents catastrophic forgetting and ultimately achieves performance exceeding the Golden Tests baseline. Figure~\ref{fig:mistake:b} reveals the adversarial dynamics: added tests (pink) represent successful attacks exposing new vulnerabilities, while removed tests (green) indicate successful defenses fixing previous bugs. After Step 38, experience replay triggers a surge in bug fixes. As training progresses, the two curves converge toward equilibrium, with the Test LLM uncovering increasingly subtle edge cases while the Code LLM repairs specific logical flaws. This dynamic balance drives synchronized improvement of both models.

\section{Analysis}

\paragraph{Generated Tests as Golden Tests.}

\begin{wraptable}[13]{r}{0.51\linewidth}
\centering
\small
\setlength{\tabcolsep}{4pt}
\vspace{-1.2\baselineskip}
\caption{Performance comparison of Code LLMs on three benchmarks using Qwen2.5-Coder-1.5B-Instruct as the base model and employing Golden Tests from various sources for RLVR. The best results are highlighted in bold, and the second best results are underlined.}
\begin{tabular}{lcccc}
\toprule
\multicolumn{1}{l}{\multirow{2}[2]{*}{\textbf{Golden Test Source}}} &
\multicolumn{1}{c}{\textbf{HE$^+$}} &
\multicolumn{1}{c}{\textbf{MBPP$^+$}} &
\multicolumn{1}{c}{\textbf{BCB}} &
\multirow{2}[2]{*}{\textbf{Avg}} \\
& \multicolumn{1}{c}{avg@32} & \multicolumn{1}{c}{avg@32} & \multicolumn{1}{c}{avg@32} & \\
\midrule
Human Annotation & 71.15 & 63.30 & \textbf{34.23} & \underline{56.23} \\
base Test LLM & 69.68 & 63.66 & 32.85 & 55.40 \\
SFT Test LLM & \underline{71.49} & \underline{64.24} & 32.93 & 56.22 \\
\rowcolor{gray!20}{\methodname Test LLM} & \textbf{71.67} & \textbf{64.42} & \underline{34.17} & \textbf{56.75} \\
\bottomrule
\end{tabular}
\label{tab:static}
\vspace{1.0\baselineskip}
\end{wraptable}

To directly assess test quality, we use tests generated by different models (Base, SFT, \methodname) as static golden tests for standard RLVR training. As shown in Table~\ref{tab:static}, the Code LLM trained with \methodname-generated tests achieves 56.75\% average accuracy, surpassing both human annotations (56.23\%) and SFT-generated tests (56.22\%). On HumanEval$^+$ and MBPP$^+$, \methodname-guided training reaches 71.67\% and 64.42\%, exceeding the human annotation baseline (71.15\% and 63.30\%). This result demonstrates that \methodname synthesizes tests of sufficient quality to replace expensive manual annotation, validating adversarial co-evolution as a scalable alternative to human supervision.

\paragraph{Training Dynamics.}

\begin{figure}[h]
\centering
\includegraphics[width=1.0\linewidth]{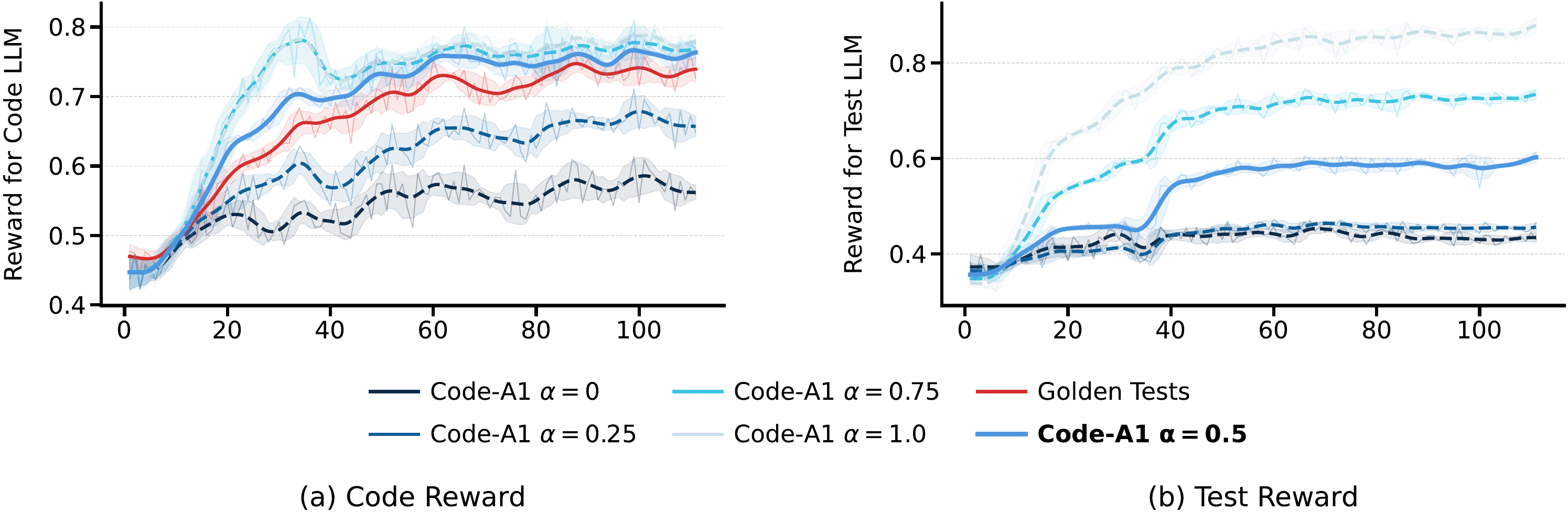}
\captionsetup{font=small, skip=6pt}
\caption{Training dynamics (Code Reward and Test Reward) over training steps for (a) Code LLM and (b) Test LLM.}
\label{fig:rewards}
\vspace{-0.8\baselineskip}
\end{figure}

Figure~\ref{fig:rewards} reveals the co-evolution process under different settings. For the Code LLM (Figure~\ref{fig:rewards}a), \methodname with $\alpha=0.5$ closely tracks the Golden Tests baseline, indicating that dynamically generated tests provide learning signals equivalent to human annotations. Extreme $\alpha$ values lead to suboptimal trajectories: $\alpha=0$ causes erratic learning due to unstable test generation, while $\alpha=1.0$ results in inflated rewards from trivial tests that fail to challenge the model. For the Test LLM (Figure~\ref{fig:rewards}b), $\alpha=0.5$ achieves the highest sustained reward, reflecting the emergence of tests that are both valid and adversarially effective. This dynamic equilibrium confirms that \methodname successfully balances the competing objectives, enabling both models to improve synchronously without manual reward calibration.

\paragraph{Test-Time Scaling.}

\begin{wraptable}{r}{0.56\linewidth}
\vspace{-1.1\baselineskip}
\centering
\small
\setlength{\tabcolsep}{4pt}
\caption{Performance of different combinations under parallel test-time scaling (see details in Appendix~\ref{appendix:tts}).}
\begin{tabular}{llcccc}
\toprule
\multicolumn{2}{c}{\textbf{Parallel TTS}} & \multirow{1}[2]{*}{\textbf{HE+}} & \multirow{1}[2]{*}{\textbf{MBPP$^+$}} & \multirow{1}[2]{*}{\textbf{BCB}} & \multirow{2}[2]{*}{\textbf{Avg}} \\
\cmidrule(r){1-2}
\textbf{Code LLM} & \textbf{Test LLM} & avg@32 & avg@32 & avg@32 & \\
\midrule
Base & / & 77.63 & 63.12 & 41.78 & 60.84\\
\methodname & / & 83.52 & 69.07 & 45.85 & 66.15\\
\midrule
Base & Base & 82.32 & 70.63 & 42.45 & 65.13\\
\methodname & Base & 84.76 & 71.69 & 45.91 & 67.45\\
\midrule
Base & \methodname & 82.93 & 70.90 & 43.63 & 65.82\\
\rowcolor{gray!20} \methodname & \methodname & 85.37 & 71.96 & 46.09 & 67.81\\
\bottomrule
\end{tabular}
\label{tab:scaling}
\vspace{-1.1\baselineskip}
\end{wraptable}

Beyond training, we evaluate whether \methodname models collaborate effectively at inference time. Table~\ref{tab:scaling} shows results under parallel test-time scaling, where the Code LLM generates multiple candidates and the Test LLM selects the best. The \methodname Test LLM demonstrates superior discriminatory power: when verifying the Base Code LLM, it achieves 65.82\% compared to 65.13\% with Base self-verification. The full combination (\methodname + \methodname) yields 67.81\%, the highest overall performance. Notably, while the Base Test LLM improves the \methodname Code LLM (66.15\% to 67.45\%), it cannot fully exploit its potential. This confirms that high-quality adversarial tests are essential for distinguishing subtle correctness differences in optimized solutions, ensuring verification capability scales with generation capability.

\paragraph{Test Difficulty \& Diversity by Case Study.}
We examine the ``threeSum'' problem (Appendix~\ref{appendix:examples}) to illustrate how test quality evolves. The Code LLM's initial solution passes both human-annotated golden tests and pre-training Test LLM outputs, yet contains a subtle bug: failure to handle duplicate triplets when moving pointers. As training progresses, the Test LLM learns to target this blind spot, generating edge cases with highly repetitive elements (e.g., [-2, 1, 1, 1, 1]) that expose the flaw. This demonstrates the curriculum-like progression: the Test LLM evolves from generic inputs to targeted adversarial attacks that probe implementation-specific vulnerabilities.

\section{Conclusion}
In this paper, we introduced \methodname, a novel reinforcement learning framework that orchestrates the adversarial co-evolution of a Code LLM and a Test LLM. By decoupling code and test generation into two models with opposing objectives, \methodname circumvents the limitations arising from self-play to avoid self-collusion and safely enables white-box testing. A Mistake Book mechanism stabilizes training by maintaining historical failures as experience replay, preventing catastrophic forgetting while providing stable baselines for reward computation. Experiments demonstrate that \methodname matches or exceeds RL with human-annotated golden tests on code generation, while producing a Test LLM that generates high-quality, bug-revealing tests more effectively than supervised fine-tuning or parameter scaling. These results suggest that adversarial co-evolution offers a scalable path beyond annotation-dependent training. Limitations and future directions are discussed in Appendix~\ref{appendix:limitation}.

\bibliography{main,custom}

\begin{thebibliography}{35}
\providecommand{\natexlab}[1]{#1}
\providecommand{\url}[1]{\texttt{#1}}
\expandafter\ifx\csname urlstyle\endcsname\relax
  \providecommand{\doi}[1]{doi: #1}\else
  \providecommand{\doi}{doi: \begingroup \urlstyle{rm}\Url}\fi

\bibitem[Altmayer~Pizzorno \& Berger(2025)Altmayer~Pizzorno and Berger]{altmayer2025coverup}
Juan Altmayer~Pizzorno and Emery~D Berger.
\newblock Coverup: Effective high coverage test generation for python.
\newblock \emph{Proceedings of the ACM on Software Engineering}, 2\penalty0 (FSE):\penalty0 2897--2919, 2025.

\bibitem[Austin et~al.(2021)Austin, Odena, Nye, Bosma, Michalewski, et~al.]{austin2021program}
Jacob Austin, Augustus Odena, Maxwell Nye, Maarten Bosma, Henryk Michalewski, et~al.
\newblock Program synthesis with large language models, August 2021.

\bibitem[Chen et~al.()Chen, Zhang, Nguyen, Zan, Lin, Lou, and Chen]{chencodet}
Bei Chen, Fengji Zhang, Anh Nguyen, Daoguang Zan, Zeqi Lin, Jian-Guang Lou, and Weizhu Chen.
\newblock Codet: Code generation with generated tests.
\newblock In \emph{The Eleventh International Conference on Learning Representations}.

\bibitem[Chen et~al.(2021)Chen, Tworek, Jun, Yuan, Pinto, et~al.]{chen2021evaluating}
Mark Chen, Jerry Tworek, Heewoo Jun, Qiming Yuan, Henrique Ponde de~Oliveira Pinto, et~al.
\newblock Evaluating large language models trained on code, July 2021.

\bibitem[Chen et~al.(2025)Chen, Wang, Zhu, Yu, Feng, Zhang, Patwary, and You]{chen2025multi}
Yixing Chen, Yiding Wang, Siqi Zhu, Haofei Yu, Tao Feng, Muhan Zhang, Mostofa Patwary, and Jiaxuan You.
\newblock Multi-agent evolve: Llm self-improve through co-evolution.
\newblock \emph{arXiv preprint arXiv:2510.23595}, 2025.

\bibitem[Cheng et~al.(2024)Cheng, Chen, Chen, Chen, Chen, Chen, Chen, Geng, Li, Li, et~al.]{cheng2024fullstack}
Yao Cheng, Jianfeng Chen, Jie Chen, Li~Chen, Liyu Chen, Wentao Chen, Zhengyu Chen, Shijie Geng, Aoyan Li, Bo~Li, et~al.
\newblock Fullstack bench: Evaluating llms as full stack coders.
\newblock \emph{arXiv preprint arXiv:2412.00535}, 2024.

\bibitem[Denison et~al.(2024)Denison, MacDiarmid, Barez, Duvenaud, Kravec, Marks, Schiefer, Soklaski, Tamkin, Kaplan, Shlegeris, Bowman, Perez, and Hubinger]{denison2024sycophancy}
Carson Denison, Monte MacDiarmid, Fazl Barez, David Duvenaud, Shauna Kravec, Samuel Marks, Nicholas Schiefer, Ryan Soklaski, Alex Tamkin, Jared Kaplan, Buck Shlegeris, Samuel~R. Bowman, Ethan Perez, and Evan Hubinger.
\newblock Sycophancy to subterfuge: Investigating reward-tampering in large language models, 2024.

\bibitem[Guo et~al.(2025)Guo, Yang, Zhang, Song, Zhang, Xu, Zhu, Ma, Wang, Bi, et~al.]{guo2025deepseek}
Daya Guo, Dejian Yang, Haowei Zhang, Junxiao Song, Ruoyu Zhang, Runxin Xu, Qihao Zhu, Shirong Ma, Peiyi Wang, Xiao Bi, et~al.
\newblock Deepseek-r1: Incentivizing reasoning capability in llms via reinforcement learning.
\newblock \emph{arXiv preprint arXiv:2501.12948}, 2025.

\bibitem[He et~al.(2025)He, Choi, Zhang, Ji, Zhou, Xu, Bercovich, Zhang, and Li]{he2025hardtests}
Zhongmou He, Yee~Man Choi, Kexun Zhang, Jiabao Ji, Junting Zhou, Dejia Xu, Ivan Bercovich, Aidan Zhang, and Lei Li.
\newblock Hardtests: Synthesizing high-quality test cases for {LLM} coding.
\newblock In \emph{NeurIPS 2025 Fourth Workshop on Deep Learning for Code}, 2025.
\newblock URL \url{https://openreview.net/forum?id=cUdqICr7aZ}.

\bibitem[Hossain \& Dwyer(2025)Hossain and Dwyer]{hossain2025togll}
Soneya~Binta Hossain and Matthew~B Dwyer.
\newblock Togll: Correct and strong test oracle generation with llms.
\newblock In \emph{2025 IEEE/ACM 47th International Conference on Software Engineering (ICSE)}, pp.\  1475--1487. IEEE, 2025.

\bibitem[Huang et~al.(2025)Huang, Zhang, Harman, Zhang, Du, and Ng]{huang2025benchmarking}
Dong Huang, Jie~M Zhang, Mark Harman, Qianru Zhang, Mingzhe Du, and See-Kiong Ng.
\newblock Benchmarking llms for unit test generation from real-world functions.
\newblock \emph{arXiv preprint arXiv:2508.00408}, 2025.

\bibitem[Hui et~al.(2024)Hui, Yang, Cui, Yang, Liu, Zhang, Liu, Zhang, Yu, Lu, et~al.]{hui2024qwen2}
Binyuan Hui, Jian Yang, Zeyu Cui, Jiaxi Yang, Dayiheng Liu, Lei Zhang, Tianyu Liu, Jiajun Zhang, Bowen Yu, Keming Lu, et~al.
\newblock Qwen2. 5-coder technical report.
\newblock \emph{arXiv preprint arXiv:2409.12186}, 2024.

\bibitem[Jain et~al.(2025)Jain, Synnaeve, and Roziere]{jain2025testgeneval}
Kush Jain, Gabriel Synnaeve, and Baptiste Roziere.
\newblock Testgeneval: A real world unit test generation and test completion benchmark.
\newblock In \emph{The Thirteenth International Conference on Learning Representations}, 2025.
\newblock URL \url{https://openreview.net/forum?id=7o6SG5gVev}.

\bibitem[Jeong et~al.()Jeong, Kim, and Park]{jeong2025ensuring}
Jaewoo Jeong, Taesoo Kim, and Sangdon Park.
\newblock Ensuring functional correctness of large code models with selective generation.
\newblock In \emph{NeurIPS 2025 Fourth Workshop on Deep Learning for Code}.

\bibitem[Kwon et~al.(2023)Kwon, Li, Zhuang, Sheng, Zheng, Yu, Gonzalez, Zhang, and Stoica]{kwon2023efficient}
Woosuk Kwon, Zhuohan Li, Siyuan Zhuang, Ying Sheng, Lianmin Zheng, Cody~Hao Yu, Joseph Gonzalez, Hao Zhang, and Ion Stoica.
\newblock Efficient memory management for large language model serving with pagedattention.
\newblock In \emph{Proceedings of the 29th symposium on operating systems principles}, pp.\  611--626, 2023.

\bibitem[Le et~al.(2022)Le, Wang, Gotmare, Savarese, and Hoi]{NEURIPS2022_8636419d}
Hung Le, Yue Wang, Akhilesh~Deepak Gotmare, Silvio Savarese, and Steven Chu~Hong Hoi.
\newblock Coderl: Mastering code generation through pretrained models and deep reinforcement learning.
\newblock In S.~Koyejo, S.~Mohamed, A.~Agarwal, D.~Belgrave, K.~Cho, and A.~Oh (eds.), \emph{Advances in Neural Information Processing Systems}, volume~35, pp.\  21314--21328. Curran Associates, Inc., 2022.
\newblock URL \url{https://proceedings.neurips.cc/paper_files/paper/2022/file/8636419dea1aa9fbd25fc4248e702da4-Paper-Conference.pdf}.

\bibitem[Liu et~al.(2023{\natexlab{a}})Liu, Zhu, Xiao, FU, Han, Wei, and Ye]{liu2023rltf}
Jiate Liu, Yiqin Zhu, Kaiwen Xiao, QIANG FU, Xiao Han, Yang Wei, and Deheng Ye.
\newblock {RLTF}: Reinforcement learning from unit test feedback.
\newblock \emph{Transactions on Machine Learning Research}, 2023{\natexlab{a}}.
\newblock ISSN 2835-8856.
\newblock URL \url{https://openreview.net/forum?id=hjYmsV6nXZ}.

\bibitem[Liu et~al.(2023{\natexlab{b}})Liu, Xia, Wang, and ZHANG]{liu2023is}
Jiawei Liu, Chunqiu~Steven Xia, Yuyao Wang, and LINGMING ZHANG.
\newblock Is your code generated by chat{GPT} really correct? rigorous evaluation of large language models for code generation.
\newblock In \emph{Thirty-seventh Conference on Neural Information Processing Systems}, 2023{\natexlab{b}}.
\newblock URL \url{https://openreview.net/forum?id=1qvx610Cu7}.

\bibitem[Lu et~al.(2025)Lu, Wen, Cheng, Ding, Xu, Guo, Wang, Chen, Jiang, and Jiang]{lu2025searchselfplay}
Hongliang Lu, Yuhang Wen, Pengyu Cheng, Ruijin Ding, Haotian Xu, Jiaqi Guo, Chutian Wang, Haonan Chen, Xiaoxi Jiang, and Guanjun Jiang.
\newblock Search self-play: Pushing the frontier of agent capability without supervision.
\newblock \emph{arXiv preprint arXiv:2510.18821}, 2025.

\bibitem[Shao et~al.(2024)Shao, Wang, Zhu, Xu, Song, et~al.]{shao2024deepseekmath}
Zhihong Shao, Peiyi Wang, Qihao Zhu, Runxin Xu, Junxiao Song, et~al.
\newblock Deepseekmath: Pushing the limits of mathematical reasoning in open language models, February 2024.

\bibitem[Shen et~al.(2023)Shen, Zhang, Chen, Zan, Geng, Fu, Zeng, Yu, Ji, Zhao, et~al.]{shen2023pangu}
Bo~Shen, Jiaxin Zhang, Taihong Chen, Daoguang Zan, Bing Geng, An~Fu, Muhan Zeng, Ailun Yu, Jichuan Ji, Jingyang Zhao, et~al.
\newblock Pangu-coder2: Boosting large language models for code with ranking feedback.
\newblock \emph{arXiv preprint arXiv:2307.14936}, 2023.

\bibitem[Sheng et~al.(2025)Sheng, Zhang, Ye, Wu, Zhang, et~al.]{sheng2025hybridflow}
Guangming Sheng, Chi Zhang, Zilingfeng Ye, Xibin Wu, Wang Zhang, et~al.
\newblock Hybridflow: A flexible and efficient rlhf framework.
\newblock In \emph{Proceedings of the Twentieth European Conference on Computer Systems}, pp.\  1279--1297, March 2025.
\newblock \doi{10.1145/3689031.3696075}.

\bibitem[Shojaee et~al.(2023)Shojaee, Jain, Tipirneni, and Reddy]{shojaee2023executionbased}
Parshin Shojaee, Aneesh Jain, Sindhu Tipirneni, and Chandan~K. Reddy.
\newblock Execution-based code generation using deep reinforcement learning.
\newblock \emph{Transactions on Machine Learning Research}, 2023.
\newblock ISSN 2835-8856.
\newblock URL \url{https://openreview.net/forum?id=0XBuaxqEcG}.

\bibitem[Tip et~al.(2025)Tip, Bell, and Sch{\"a}fer]{tip2025llmorpheus}
Frank Tip, Jonathan Bell, and Max Sch{\"a}fer.
\newblock Llmorpheus: Mutation testing using large language models.
\newblock \emph{IEEE Transactions on Software Engineering}, 2025.

\bibitem[Wang et~al.(2025)Wang, Yang, Tian, Shen, and Wang]{wang2025cure}
Yinjie Wang, Ling Yang, Ye~Tian, Ke~Shen, and Mengdi Wang.
\newblock {CURE}: Co-evolving coders and unit testers via reinforcement learning.
\newblock In \emph{The Thirty-ninth Annual Conference on Neural Information Processing Systems}, 2025.
\newblock URL \url{https://openreview.net/forum?id=wPdBe9zxNr}.

\bibitem[Xu et~al.(2025)Xu, Liu, Yin, Zhou, and Poovendran]{xu2025kodcode}
Zhangchen Xu, Yang Liu, Yueqin Yin, Mingyuan Zhou, and Radha Poovendran.
\newblock Kodcode: A diverse, challenging, and verifiable synthetic dataset for coding.
\newblock \emph{arXiv preprint arXiv:2503.02951}, 2025.

\bibitem[Yang et~al.(2025)Yang, Lieret, Jimenez, Wettig, Khandpur, Zhang, Hui, Press, Schmidt, and Yang]{yang2025swesmith}
John Yang, Kilian Lieret, Carlos~E Jimenez, Alexander Wettig, Kabir Khandpur, Yanzhe Zhang, Binyuan Hui, Ofir Press, Ludwig Schmidt, and Diyi Yang.
\newblock {SWE}-smith: Scaling data for software engineering agents.
\newblock In \emph{The Thirty-ninth Annual Conference on Neural Information Processing Systems Datasets and Benchmarks Track}, 2025.
\newblock URL \url{https://openreview.net/forum?id=63iVrXc8cC}.

\bibitem[Yu et~al.(2025)Yu, Zhang, Zhu, Yuan, Zuo, et~al.]{yu2025dapo}
Qiying Yu, Zheng Zhang, Ruofei Zhu, Yufeng Yuan, Xiaochen Zuo, et~al.
\newblock Dapo: An open-source llm reinforcement learning system at scale.
\newblock In \emph{The Thirty-Ninth Annual Conference on Neural Information Processing Systems}, October 2025.

\bibitem[Zeng et~al.(2025)Zeng, Jiang, Wang, Nie, Chen, and Chen]{zeng2025acecoder}
Huaye Zeng, Dongfu Jiang, Haozhe Wang, Ping Nie, Xiaotong Chen, and Wenhu Chen.
\newblock Acecoder: Acing coder rl via automated test-case synthesis.
\newblock In \emph{Proceedings of the 63rd Annual Meeting of the Association for Computational Linguistics (Volume 1: Long Papers)}, pp.\  12023--12040, 2025.

\bibitem[Zhan et~al.(2025)Zhan, Li, Wang, Qu, Liu, Shao, Wong, and Cheng]{zhan2025exgrpo}
Runzhe Zhan, Yafu Li, Zhi Wang, Xiaoye Qu, Dongrui Liu, Jing Shao, Derek~F Wong, and Yu~Cheng.
\newblock Exgrpo: Learning to reason from experience.
\newblock \emph{arXiv preprint arXiv:2510.02245}, 2025.

\bibitem[Zhang et~al.(2024{\natexlab{a}})Zhang, Wu, Yang, Shu, Xiao, Kong, and Sang]{zhang2024o1}
Yuxiang Zhang, Shangxi Wu, Yuqi Yang, Jiangming Shu, Jinlin Xiao, Chao Kong, and Jitao Sang.
\newblock o1-coder: an o1 replication for coding.
\newblock \emph{arXiv preprint arXiv:2412.00154}, 2024{\natexlab{a}}.

\bibitem[Zhang et~al.(2024{\natexlab{b}})Zhang, Wu, Yang, Shu, Xiao, Kong, and Sang]{zhang2024o1codero1replicationcoding}
Yuxiang Zhang, Shangxi Wu, Yuqi Yang, Jiangming Shu, Jinlin Xiao, Chao Kong, and Jitao Sang.
\newblock O1-coder: An o1 replication for coding, 2024{\natexlab{b}}.
\newblock URL \url{https://arxiv.org/abs/2412.00154}.

\bibitem[Zhao et~al.(2025)Zhao, Wu, Yue, Wu, Xu, Yue, Lin, Wang, Wu, Zheng, and Huang]{zhao2025absolute}
Andrew Zhao, Yiran Wu, Yang Yue, Tong Wu, Quentin Xu, Yang Yue, Matthieu Lin, Shenzhi Wang, Qingyun Wu, Zilong Zheng, and Gao Huang.
\newblock Absolute zero: Reinforced self-play reasoning with zero data.
\newblock In \emph{The Thirty-ninth Annual Conference on Neural Information Processing Systems}, 2025.
\newblock URL \url{https://openreview.net/forum?id=neZSGqhxDa}.

\bibitem[Zhao et~al.(2023)Zhao, Gu, Varma, Luo, Huang, et~al.]{zhao2023pytorch}
Yanli Zhao, Andrew Gu, Rohan Varma, Liang Luo, Chien-Chin Huang, et~al.
\newblock Pytorch fsdp: Experiences on scaling fully sharded data parallel, September 2023.

\bibitem[Zhuo et~al.(2025)Zhuo, Chien, Chim, Hu, Yu, Widyasari, Yusuf, Zhan, He, Paul, Brunner, GONG, Hoang, Zebaze, Hong, Li, Kaddour, Xu, Zhang, Yadav, Jain, Gu, Cheng, Liu, Liu, Wang, Lo, Hui, Muennighoff, Fried, Du, de~Vries, and Werra]{zhuo2025bigcodebench}
Terry~Yue Zhuo, Vu~Minh Chien, Jenny Chim, Han Hu, Wenhao Yu, Ratnadira Widyasari, Imam Nur~Bani Yusuf, Haolan Zhan, Junda He, Indraneil Paul, Simon Brunner, Chen GONG, James Hoang, Armel~Randy Zebaze, Xiaoheng Hong, Wen-Ding Li, Jean Kaddour, Ming Xu, Zhihan Zhang, Prateek Yadav, Naman Jain, Alex Gu, Zhoujun Cheng, Jiawei Liu, Qian Liu, Zijian Wang, David Lo, Binyuan Hui, Niklas Muennighoff, Daniel Fried, Xiaoning Du, Harm de~Vries, and Leandro~Von Werra.
\newblock Bigcodebench: Benchmarking code generation with diverse function calls and complex instructions.
\newblock In \emph{The Thirteenth International Conference on Learning Representations}, 2025.
\newblock URL \url{https://openreview.net/forum?id=YrycTjllL0}.

\end{thebibliography}
\bibliographystyle{main}

\clearpage
\tableofcontents
\clearpage
\appendix

\section{Experimental Details.}
\label{appendix:experimental_details}
\subsection{Prompt Design}
\begin{AIbox}{Code LLM}
\textless\textbar im\_start\textbar\textgreater system\newline
You are a helpful code completion assistant.\newline
\textless\textbar im\_end\textbar\textgreater\newline
\textless\textbar im\_start\textbar\textgreater user\newline
Given the following Question, complete the function. Output the complete function inside \verb|```python ... ```| code block, and do not output anything else.\newline
Question:\newline
\{question\}\newline
\textless\textbar im\_end\textbar\textgreater\newline
\textless\textbar im\_start\textbar\textgreater assistant
\end{AIbox}

\begin{AIbox}{Test LLM}
\textless\textbar im\_start\textbar\textgreater system\newline
You are a helpful test case generation assistant.\newline
\textless\textbar im\_end\textbar\textgreater\newline
\textless\textbar im\_start\textbar\textgreater user\newline
\# Role\newline
You are specializing in finding specific inputs that cause \verb|`|Buggy Code\verb|`| to behave differently from the requirements (\verb|`|Question\verb|`|).\newline
\newline
\# Task\newline
Generate 8 assertion-based test cases to detect bugs for the function in \verb|`|Buggy Code\verb|`| according to the \verb|`|Question\verb|`|. \newline
\newline
\# Strategy\newline
1. Attack Logic Gaps: Analyze where the \verb|`|Buggy Code\verb|`| logic might be too simple compared to the \verb|`|Question\verb|`|. Construct input \verb|`|parameters\verb|`| that hit these blind spots (e.g., missing constraints, misinterpreted rules, over-simplified logic).\newline
2. Prioritize Complexity: Prefer complex input \verb|`|parameters\verb|`| (e.g., boundary values, nested loops, compound conditions, rare branches) over simple ones. Ensure every logical branch is stressed and every potential issue is covered.\newline
3. Zero Redundancy: Do not brute-force generating trivial or repetitive tests. Only the first few generated tests will be evaluated, so quality and ordering matter more than quantity.\newline
\newline
\# Context\newline
Question:\newline
\{question\}\newline
\newline
Buggy Code:\newline
\verb|```python|\newline
\{generated\_code\}\newline
\verb|```|\newline
\newline
\# Output Format\newline
Output ALL the assert statements inside ONE \verb|```python ... ```| code block.\newline
Format: \verb|assert function_name(parameters) == answer|\newline
\textless\textbar im\_end\textbar\textgreater\newline
\textless\textbar im\_start\textbar\textgreater assistant
\end{AIbox}

\subsection{Sandbox Design}
Given that code training and validation heavily rely on stable sandbox environments, we deployed a sandbox service using Sandbox Fusion~\citep{cheng2024fullstack}. This service organizes code and tests for batch execution within the sandbox, retrieving execution results via the sandbox's stdout. The code template executed in the sandbox is as follows:

\textbf{Template 1: Validating and Executing in Training Phase.}
Template 1 is designed for efficient interaction during the training phase. It employs a high-reusability batch execution strategy, compressing all sandbox interaction requests for a single training step into one, thereby significantly reducing high concurrency pressure on the sandbox. The process follows a ``calibrate-then-evaluate" mechanism: first, the Ground Truth code is introduced as an Oracle to execute and verify the test cases generated by the Test LLM, filtering out invalid cases that are malformed or inconsistent with standard outputs; subsequently, in a cleaned context, the validated new tests are combined with historical tests from the Mistake Book to comprehensively evaluate the code generated by the Code LLM, thus collecting all necessary data for reward computation in a single execution.
\begin{lstlisting}[language=Python]
<GLOBAL_IMPORTS>
valid_asserts_list = []
try:
    <GROUND_TRUTH_CODE>
    for n in range(<NUM_SAMPLES>):
        invalid_count = <INVALID_COUNT>
        valid_asserts = []
        seen = set()
        gen_total = 0
        for stmt, call, ans in zip(<NORMALIZED_STMTS>, <CALLS>, <ANSWERS>):
            try:
                _r = <FUNCTION_CALL>
                stmt = stmt.replace('__TO_BE_FILLED__', repr(_r))
                if stmt in seen:
                    invalid_count += 1
                    valid_asserts.append(None)
                else:
                    seen.add(stmt)
                    valid_asserts.append({'val': _r, 'stmt': stmt})
                    print('__GEN_START__<TEST_ID>_' + str(gen_total) + ':' + repr(stmt), flush=True)
                    gen_total += 1
                    if (_r) != (<EXPECTED_ANSWER>):
                        invalid_count += 1
            except Exception:
                invalid_count += 1
                valid_asserts.append(None)
        print('__INVALID_TEST__<TEST_ID>:' + str(invalid_count), flush=True)
        valid_asserts_list.append(valid_asserts)
except Exception:
    pass

for gt_fn in <GT_FUNCTION_NAMES>:
    try:
        del <GT_FN>
    except Exception:
        pass

try:
    <GENERATED_CODE>
    print('__CODE_VALID__<CODE_ID>', flush=True)
    
    for idx, test_item in enumerate(<GT_TESTCASE_LIST>):
        try:
            <GT_TEST_CODE>
            print('__GT_PASS__<CODE_ID>_' + str(idx), flush=True)
        except Exception as e:
            print('__GT_FAIL__<CODE_ID>_' + str(idx) + ':' + repr(e), flush=True)
            
    for idx, test_item in enumerate(<ATTACK_TESTCASE_LIST>):
        print('__ATTACK_START__<CODE_ID>_' + str(idx) + ':' + repr(repr(test_item)), flush=True)
        try:
            <ATTACK_TEST_CODE>
            print('__ATTACK_PASS__<CODE_ID>_' + str(idx), flush=True)
        except Exception as e:
            print('__ATTACK_FAIL__<CODE_ID>_' + str(idx) + ':' + repr(e), flush=True)
            
    for n in range(<NUM_SAMPLES>):
        valid_asserts = valid_asserts_list[n]
        gen_total = 0
        for j, call in enumerate(<CALLS_LIST>):
            spec = valid_asserts[j]
            if spec is not None:
                try:
                    assert <FUNCTION_CALL> == spec['val']
                    print('__GEN_PASS__<TEST_ID>_' + str(gen_total), flush=True)
                except Exception as e:
                    print('__GEN_FAIL__<TEST_ID>_' + str(gen_total) + ':' + repr(e), flush=True)
                gen_total += 1
except Exception:
    pass
\end{lstlisting}

\textbf{Template 2: Evaluation in Validation Phase.}
Template 2 is explicitly designed for the inference evaluation phase of the Code LLM, aiming to efficiently complete the batch validation of multiple sampled solutions for the same question via a single sandbox request. The script independently encapsulates each generated function body using a signal-based timeout protection mechanism (run\_safe) to prevent infinite loops in individual samples from interrupting the overall evaluation task. By sequentially calling the standard test harness within the same context, it rapidly verifies the functional correctness of large-scale candidate code; this design significantly enhances the concurrency efficiency for calculating metrics such as avg@k while ensuring execution safety.
\begin{lstlisting}[language=Python]
<IMPORTS>
import signal

class TO(Exception): pass
def handler(s, f): raise TO()

def run_safe(func):
    try:
        signal.signal(signal.SIGALRM, handler)
        signal.alarm(<TIMEOUT>)
        func()
        signal.alarm(0)
        print('__PASS__', flush=True)
    except TO:
        print('__TIMEOUT__', flush=True)
    except:
        signal.alarm(0)
        print('__FAIL__', flush=True)

<TEST_HARNESS_CODE>

def c_0():
    <GENERATED_CODE_0>
    check(<ENTRY_POINT>)

def c_1():
    <GENERATED_CODE_1>
    check(<ENTRY_POINT>)

# ... (repeated for batch size) ...

run_safe(c_0)
run_safe(c_1)
# ... (repeated for batch size) ...
\end{lstlisting}

\subsection{Mistake Book Design}

\label{appendix:mistake}

The Mistake Book is implemented as a global variable stored in JSON format, designed to stabilize adversarial training through a persistent experience replay mechanism. The data structure utilizes a key-value pair format: the Key serves as the unique identifier for each question (question\_id), while the Value is a list of dictionaries containing specific failed test cases and their corresponding failure frequencies (frequency). 
\paragraph{Usage.} For each question in the current training step, the system first queries the Mistake Book. If historical failure records exist, these challenging ``mistakes" are retrieved and appended to the list of tests newly generated by the Test LLM, to be executed together in the sandbox. 
\paragraph{Update.} The update mechanism operates at the level of individual assertion test points. If the code fails a specific test, that test is added to the Mistake Book (or its frequency is incremented). Conversely, if the code successfully passes a known test from the Mistake Book, its frequency is decremented. When the frequency drops to zero, the test case is considered ``mastered" and is removed from the Mistake Book. 
\paragraph{Persistence.} To support training interruption and resumption, the system serializes and saves the current global Mistake Book to a JSON file after every step. This ensures that historical adversarial experiences can be seamlessly reloaded when resuming training.
\begin{lstlisting}
{
    "Apps_1564_I": [
      {
        "testcase": "assert number_of_characters(5) == 2",
        "frequency": 4
      },
      {
        "testcase": "assert number_of_characters(10) == 3",
        "frequency": 5
      },
      ...
    ],
    "Leetcode_17190_I": [
      {
        "testcase": "assert can_rearrange_to_palindrome(\"aaa\", 1) == False",
        "frequency": 10
      },
      {
        "testcase": "assert can_rearrange_to_palindrome(\"aa\", 1) == False",
        "frequency": 12
      },
      ...
    ]
    ...
}
\end{lstlisting}

\subsection{Training Details}

\subsubsection{General RL Details}
Our RL experiments are conducted using verl~\citep{sheng2025hybridflow} as the training framework, with vLLM~\citep{kwon2023efficient} serving as the inference engine and FSDP~\citep{zhao2023pytorch} as the training backend. During the rollout phase, we sequentially sample from the Code LLM and Test LLM (if exists), submitting post-processed and concatenated responses to a sandbox server subject to a strict 10-second execution timeout to prevent training bottlenecks. To ensure robustness against sandbox instability, we implement a bidirectional monitoring mechanism: a supervisor script performs 30-second health checks to auto-restart Docker services upon anomalies, while execution failures trigger a retry protocol with a 60-second backoff and a maximum of five attempts. Samples that remain unresponsive are assigned a zero reward, ensuring a zero advantage estimate—since grouped samples share identical rewards—to prevent erroneous policy updates. Finally, we compute distinct rewards for each model based on execution outcomes and perform asynchronous parameter updates. 
All experiments are performed on 8 NVIDIA H20 GPUs with CUDA~12.8. The detailed RL hyperparameters are summarized in Table~\ref{tab:rl_params}.

\begin{table}[!t]
\centering
\small
\setlength{\tabcolsep}{4pt}
\caption{Detailed hyperparameters for reinforcement learning experiments.}
\begin{tabular}{lccc}
\toprule
\textbf{Hyperparameter} & \textbf{Golden Tests (GT)} & \textbf{Self-Play (SP)} & \textbf{\methodname} \\
\midrule
\rowcolor{lightgray!70}\multicolumn{4}{l}{\textit{Experiment Setup \& Resources}} \\
Test LLM Source & - & Self-Play (Shared) & Independent Model \\
Reward Manager & \texttt{acg\_gt} & \texttt{acg\_llm} & \texttt{acg\_llm} \\
Reward Alpha ($\alpha$) & - & 0.5 & 0.5 \\
n\_gpus\_per\_node & 8 & 8 & 4 \\
\midrule
\rowcolor{lightgray!70}\multicolumn{4}{l}{\textit{Trainer}} \\
total\_epochs & 3 & 3 & 3 \\
\midrule
\rowcolor{lightgray!70}\multicolumn{4}{l}{\textit{Algorithm}} \\
adv\_estimator & grpo & grpo & grpo \\
kl\_coef & 0.001 & 0.001 & 0.001 \\
\midrule
\rowcolor{lightgray!70}\multicolumn{4}{l}{\textit{Data}} \\
train\_batch\_size & 256 & 256 & 256 \\
max\_prompt\_length & 4096 & 4096 & 4096 \\
max\_response\_length & 4096 & 4096 & 4096 \\
\midrule
\rowcolor{lightgray!70}\multicolumn{4}{l}{\textit{Model (Actor)}} \\
model.use\_remove\_padding & true & true & true \\
model.enable\_gradient\_checkpointing & true & true & true \\
actor.ppo\_mini\_batch\_size & 128 & 128 & 128 \\
actor.use\_dynamic\_bsz & true & true & true \\
actor.ppo\_max\_token\_len\_per\_gpu & 32768 & 32768 & 32768 \\
actor.clip\_ratio\_low & 0.20 & 0.20 & 0.20 \\
actor.clip\_ratio\_high & 0.28 & 0.28 & 0.28 \\
actor.optim.lr & 1e-6 & 1e-6 & 1e-6 \\
actor.weight\_decay & 0.1 & 0.1 & 0.1 \\
\midrule
\rowcolor{lightgray!70}\multicolumn{4}{l}{\textit{Rollout}} \\
rollout.name & vllm & vllm & vllm \\
rollout.mode & sync & sync & sync \\
rollout.tensor\_model\_parallel\_size & 1 & 1 & 1 \\
rollout.gpu\_memory\_utilization & 0.8 & 0.8 & 0.8 \\
rollout.n & 8 & 8 & 8 \\
rollout.temperature & 1.0 & 1.0 & 1.0 \\
rollout.top\_p & 1.0 & 1.0 & 1.0 \\
rollout.top\_k & -1 & -1 & -1 \\
\midrule
\rowcolor{lightgray!70}\multicolumn{4}{l}{\textit{Reference Model}} \\
ref.log\_prob\_use\_dynamic\_bsz & true & true & true \\
ref.log\_prob\_max\_token\_len\_per\_gpu & 32768 & 32768 & 32768 \\
\bottomrule
\end{tabular}
\label{tab:rl_params}
\end{table}

\subsubsection{Asymmetric Sampling and $\textsc{TopVar}$}
\label{appendix:topvar}

For \methodname, in the asymmetric sampling phase, we employ the $\textsc{TopVar}$ algorithm to filter for the most informative groups based on reward variance.

Formally, let $R_{m,n}$ denote the reward for the $n$-th test suite generated conditioned on the $m$-th candidate solution $\hat{C}_m$. We calculate the intra-group reward standard deviation $\sigma_m$ for each $m \in \{1, \dots, M\}$:
\begin{equation}
\sigma_m = \sqrt{\frac{1}{N} \sum_{n=1}^{N} (R_{m,n} - \bar{R}_m)^2}, \quad \text{where } \bar{R}_m = \frac{1}{N} \sum_{n=1}^{N} R_{m,n}
\end{equation}
We then derive the index set $\mathcal{I}^*$ corresponding to the groups with the highest variance:
\begin{equation}
\mathcal{I}^* = \operatorname*{arg\,top-\ell}_{m \in \{1, \dots, M\}} (\sigma_m)
\end{equation}
The gradient update for the Test LLM is strictly restricted to these selected groups to maximize training efficiency:
\begin{equation}
\nabla \mathcal{J}_{\text{Test}} \approx \frac{1}{\ell \times N} \sum_{m \in \mathcal{I}^*} \sum_{n=1}^{N} \hat{A}_{m,n} \nabla_\theta \log \pi_{\theta}(\hat{T}_{m,n} | Q, \hat{C}_m)
\end{equation}
\subsubsection{Self-Play Details}
\label{appendix:selfplay}
The Self-Play architecture employs a unified model to fulfill both Code LLM and Test LLM roles. To mitigate the risk of reward hacking during the interaction of these dual roles—such as the model generating trivial tests to game the reward system—strict input isolation is enforced during the test generation phase: the model is restricted to referencing only the question description and is prohibited from accessing the generated candidate code. Unlike the targeted adversarial generation in \methodname, this means that all code samples for a given question are evaluated against the same set of generic, question-based tests. This lack of targeted adaptation limits the diversity and complexity of tests in exposing specific logical flaws, making the overall training process structurally similar to traditional Golden Tests-based reinforcement learning, distinguishing itself primarily by the source of the tests and the need for self-referential policy updates.

Unlike the rollout design in \methodname, during Self-Play, the model generates $M$ candidate code solutions and $N$ sets of candidate unit test suites for each programming question. In the experiments, $M=8$ and $N=8$ were set. Finally, all generated code is cross-run with all generated unit tests, and subsequent steps are completed based on the execution results following the \methodname design.

\subsubsection{SFT Details}
\label{appendix:sft}
To strictly quantify the performance gains of RL training compared to Supervised Fine-Tuning (SFT) for the Test LLM under identical data distributions, we constructed the SFT dataset based on the same training source used in the RL phase. To facilitate convergence, we adopted a simplified prompt structure. Specifically, the input Buggy Code consists of pre-sampled solution candidates that failed to pass the complete set of Golden Tests, while the output labels utilize the corresponding ground truth Golden Tests. This design intends to guide the model, via imitation learning, to master the generation of standard test cases given flawed code, thereby establishing a robust SFT baseline for the comparative reinforcement learning experiments.

\begin{AIbox}{Prompt for Test LLM in SFT}
\textless\textbar im\_start\textbar\textgreater system\newline
You are a helpful test case generation assistant.\newline
\textless\textbar im\_end\textbar\textgreater\newline
\textless\textbar im\_start\textbar\textgreater user\newline
Given the following Question, generate 8 assert-based test cases to detect bugs in the following python function Buggy Code. Each test case must be in the format: \verb|assert function_name(parameters) == answer'|, and split by \verb|\n|. Output all assert statements inside \verb|```python ... ```| code block, and do not output anything else.\newline
\newline
Question:\newline
\{question\}\newline
\newline
Buggy Code:\newline
\verb|```python|\newline
\{buggy\_code\}\newline
\verb|```|\newline
\textless\textbar im\_end\textbar\textgreater\newline
\textless\textbar im\_start\textbar\textgreater assistant
\end{AIbox}

\subsection{Test-time Scaling Design}
\label{appendix:tts}
To evaluate the collaborative performance of the Code LLM and Test LLM during the inference phase, we implemented a parallel scaling strategy based on the Best-of-N (BoN) selection mechanism based on Qwen2.5-Coder-3B-Instruct. For each question, the Code LLM samples $M=16$ candidate solutions, and the Test LLM generates $N=16$ corresponding unit test suites. Unlike the training phase, the Test LLM here generates test suites based solely on the question description, blind to the candidate solutions, serving as an independent verifier. We execute a full interaction matrix where every candidate solution is tested against every generated test suite. The candidate solution that passes the highest number of unique test cases across all suites is selected as the final answer. 

In Table~\ref{tab:scaling}, the symbol `/' in the \textbf{Test LLM} column denotes the baseline setting without scaling methods, reporting the direct avg@32 results.

\section{Examples.}
\label{appendix:examples}
\subsection{Question Example}
\begin{lstlisting}[language=Python]
from typing import List
def threeSum(nums: List[int], target: int) -> List[List[int]]:
    """
	Write a function that receives an array of integers and a target integer. Your task is to find all unique triplets in the array which gives the sum of the target integer. The solution set must not contain duplicate triplets. 
    **Example:**
    - Input: `nums = [-1, 0, 1, 2, -1, -4], target = 0`
    - Output: `[[-1, -1, 2], [-1, 0, 1]]`
    """
\end{lstlisting}

\subsection{Ground Truth Example}
\begin{lstlisting}[language=Python]
from typing import List
def threeSum(nums: List[int], target: int) -> List[List[int]]:
    nums.sort()  # Sort the array to help avoid duplicates and use two-pointer strategy
    result = []
    
    for i in range(len(nums) - 2):
        if i > 0 and nums[i] == nums[i - 1]:  # Skip the same element to avoid duplicates
            continue
        
        left, right = i + 1, len(nums) - 1
        while left < right:
            current_sum = nums[i] + nums[left] + nums[right]
            if current_sum == target:
                result.append([nums[i], nums[left], nums[right]])
                while left < right and nums[left] == nums[left + 1]:  # Skip duplicates
                    left += 1
                while left < right and nums[right] == nums[right - 1]:  # Skip duplicates
                    right -= 1
                left += 1
                right -= 1
            elif current_sum < target:
                left += 1
            else:
                right -= 1
    
    return result
\end{lstlisting}

\subsection{Golden Tests Example.}
\begin{lstlisting}[language=Python]
assert threeSum([-1, 0, 1, 2, -1, -4], 0) == [[-1, -1, 2], [-1, 0, 1]]
assert threeSum([1, 2, -2, -1], 1) == [[-2, 1, 2]]
assert threeSum([0, 0, 0], 0) == [[0, 0, 0]]
assert threeSum([], 0) == []
assert threeSum([3, 0, -2, -1, 1, 2], 0) == [[-2, -1, 3], [-2, 0, 2], [-1, 0, 1]]
assert threeSum([-1, 0, 1, 2, -1, -4, -2, -3, 3, 0, 4], 0) == [[-4, 0, 4], [-4, 1, 3], [-3, -1, 4], [-3, 0, 3], [-3, 1, 2], [-2, -1, 3], [-2, 0, 2], [-1, -1, 2], [-1, 0, 1]]
assert threeSum([1, 2, 3, 4], 10) == []
assert threeSum([-1, 2, 1, 4], 8) == []
\end{lstlisting}

\subsection{Code LLM Response Example}
\begin{lstlisting}[language=Python]
```python
def threeSum(nums, target):
    nums.sort()
    res = []
    n = len(nums)
    for i in range(n):
        if i > 0 and nums[i] == nums[i-1]:
            continue
        l, r = i + 1, n - 1
        while l < r:
            s = nums[i] + nums[l] + nums[r]
            if s == target:
                res.append([nums[i], nums[l], nums[r]])
                l += 1
                r -= 1
            elif s < target:
                l += 1
            else:
                r -= 1
    return res
```
\end{lstlisting}

\subsection{Test LLM Response Example}
\textbf{Before Training}
\begin{lstlisting}[language=Python]
```python
assert threeSum([-1, 0, 1], 0) == [[-1, 0, 1]]
assert threeSum([], 0) == []
assert threeSum([0, 0, 0], 0) == [[0, 0, 0]]
assert threeSum([0, 0, 0, 0], 0) == [[0, 0, 0]]
assert threeSum([-1, 0, 1, 2, -1, -4], 0) == [[-1, -1, 2], [-1, 0, 1]]
assert threeSum([1, 2, 3, 4], 0) == []
assert threeSum([-1, -1, 0, 1], 0) == [[-1, 0, 1]]
assert threeSum([3, 0, -2, -1, 1, 2], 0) == [[-2, -1, 3], [-2, 0, 2], [-1, 0, 1]]
```
\end{lstlisting}

\textbf{After Training}
\begin{lstlisting}[language=Python]
```python
assert threeSum([0, 0, 0, 0, 0], 0) == [[0, 0, 0]]
assert threeSum([-2, 1, 1, 1, 1], 0) == [[-2, 1, 1]]
assert threeSum([-2, 0, 0, 2, 2], 0) == [[-2, 0, 2]]
assert threeSum([-1, -1, -1, 2, 2], 0) == [[-1, -1, 2]]
assert threeSum([-4, 2, 2, 2, 2], 0) == [[-4, 2, 2]]
assert threeSum([-6, 3, 3, 3, 3, 3], 0) == [[-6, 3, 3]]
assert threeSum([-2, 1, 1, 1, 1, 0, 2], 0) == [[-2, 0, 2], [-2, 1, 1]]
assert threeSum([-100, 50, 50, 50, 50], 0) == [[-100, 50, 50]]
```
\end{lstlisting}

\section{Performance Comparison with CURE}

\label{appendix:cure}

We compare \methodname against ReasonFlux-Coder from CURE~\citep{wang2025cure}, a self-play RL framework for code generation. Note that direct comparison is confounded by base model differences: \methodname uses Qwen2.5-Coder-Instruct while ReasonFlux-Coder uses general Qwen2.5-Instruct or Long-CoT Qwen3 models. Nevertheless, cross-scale analysis reveals the efficiency advantages of adversarial co-evolution over self-play.

\begin{table*}[h]
  \centering
  \small
  \caption{\textbf{Comparison with ReasonFlux-Coder (CURE).} \methodname achieves comparable or superior performance with significantly fewer parameters. $\dagger$: Long-CoT base model.}
  \setlength{\tabcolsep}{6pt}
    \resizebox{\columnwidth}{!}{
    \begin{tabular}{lrccccccc}
    \toprule
    \multirow{2}[2]{*}{\textbf{Model}} & \multirow{2}[2]{*}{\textbf{Params}} & \multicolumn{1}{c}{\textbf{HumanEval$^+$}} & \multicolumn{1}{c}{\textbf{MBPP$^+$}} & \multicolumn{1}{c}{\textbf{BigCodeBench}} & \multirow{2}[2]{*}{\textbf{Avg}} & \multicolumn{2}{c}{\textbf{UnLeakedTestBench}} & \multirow{2}[2]{*}{\textbf{Mul}} \\
    \cmidrule(lr){7-8}
          &       & \multicolumn{1}{c}{avg@32} & \multicolumn{1}{c}{avg@32} & \multicolumn{1}{c}{avg@32} &       & \multicolumn{1}{c}{pass@5} & \multicolumn{1}{c}{mut@5} &  \\
    \midrule
    ReasonFlux-Coder-14B & 14B & 78.73 & 72.93 & 52.50 & 68.05 & 32.18 & 53.27 & 17.14 \\
    \rowcolor{gray!20} \methodname-7B & 7B \textcolor{Green}{(2$\times\!\downarrow$)} & \textbf{85.21} & \textbf{74.50} & 52.46 & \textbf{70.72} & \textbf{37.15} & 53.14 & \textbf{19.74} \\
    \midrule
    ReasonFlux-Coder-7B & 7B & 76.62 & 69.88 & 46.49 & 64.33 & 23.65 & 46.28 & 10.95 \\
    \rowcolor{gray!20} \methodname-3B & 3B \textcolor{Green}{(2.3$\times\!\downarrow$)} & \textbf{83.52} & 69.07 & 45.85 & \textbf{66.15} & \textbf{30.86} & \textbf{49.56} & \textbf{15.29} \\
    \midrule
    ReasonFlux-Coder-4B$^\dagger$ & 4B & 61.81 & \textbf{66.08} & \textbf{46.99} & \textbf{58.29} & \textbf{36.90} & \textbf{57.73} & \textbf{21.30} \\
    \rowcolor{gray!20} \methodname-1.5B & 1.5B \textcolor{Green}{(2.7$\times\!\downarrow$)} & \textbf{72.69} & 63.33 & 34.82 & 56.95 & 27.05 & 26.41 & 7.14 \\
    \bottomrule
    \end{tabular}%
    }
  \label{tab:cure}%
\end{table*}

\paragraph{Parameter Efficiency.} Table~\ref{tab:cure} shows that \methodname achieves strong performance with significantly fewer parameters. The 3B \methodname model outperforms ReasonFlux-Coder-7B on both code generation (66.15\% vs 64.33\% Avg) and test generation (15.29 vs 10.95 Mul). Similarly, \methodname-7B surpasses ReasonFlux-Coder-14B (70.72\% vs 68.05\% Avg; 19.74 vs 17.14 Mul). This suggests that decoupled adversarial co-evolution enables more efficient capability acquisition than unified self-play, consistent with our analysis that architectural separation provides fundamental advantages over single-model approaches.

\paragraph{Effect of Base Model Architecture.} The comparison between \methodname-1.5B and ReasonFlux-Coder-4B reveals an interesting asymmetry. On code generation, \methodname-1.5B approaches the 4B model despite 3$\times$ fewer parameters (56.95\% vs 58.29\% Avg). However, a substantial gap remains on test generation (7.14 vs 21.30 Mul). We attribute this to ReasonFlux-Coder-4B's Long-CoT base model, which provides inherent advantages for reasoning in test synthesis. This suggests a promising direction: combining \methodname's adversarial framework with Long-CoT architectures may yield further gains.

\section{Training Pipeline of \methodname}

\begin{figure*}[h]
    \centering
    \includegraphics[width=1.0\textwidth]{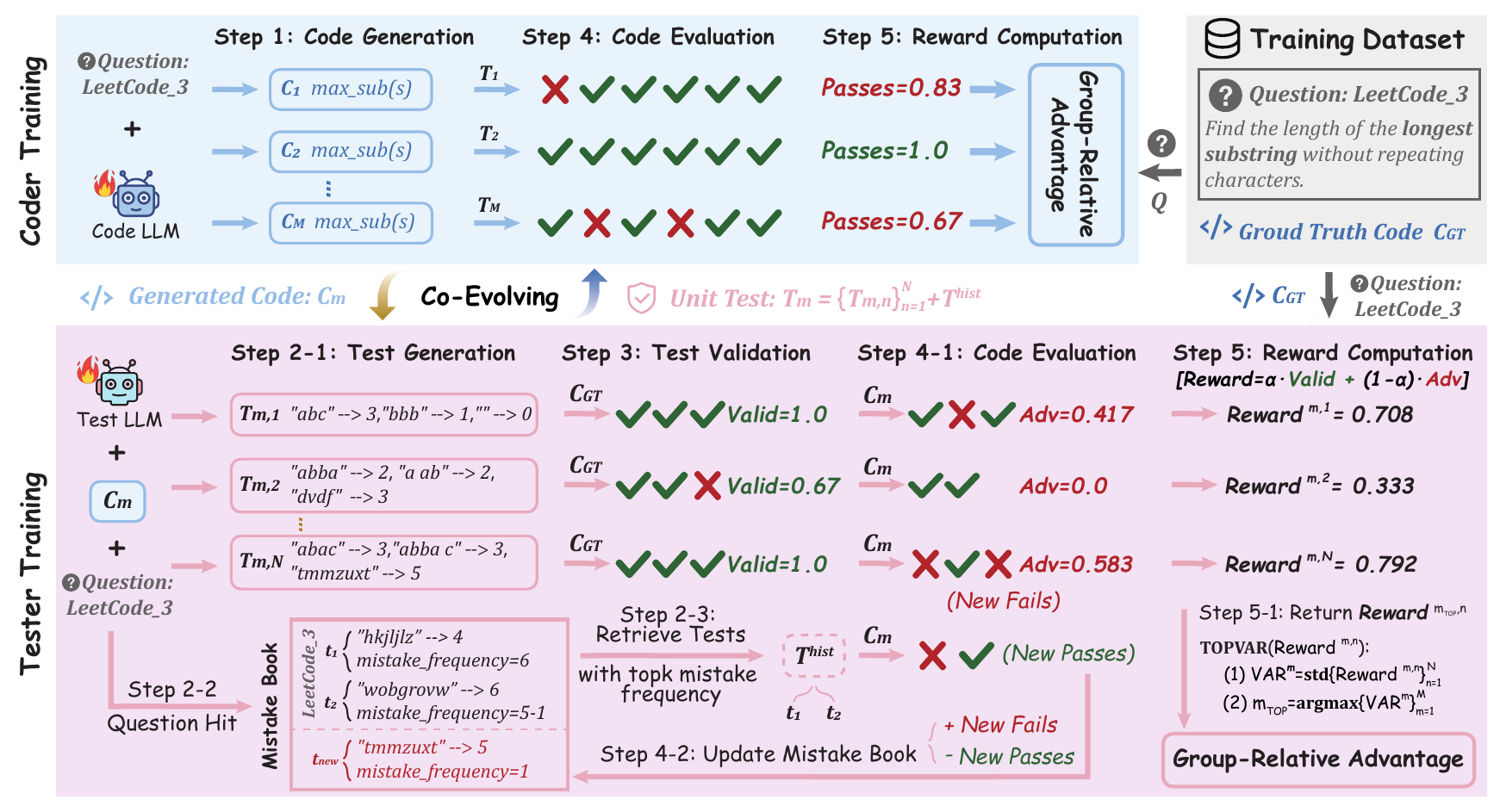}
    \caption{\textbf{Detailed pipeline of \methodname.} Top: Code LLM training with code generation (Step 1), evaluation against combined tests (Step 4), and reward computation via group-relative advantage (Step 5). Bottom: Test LLM training with white-box test generation (Step 2-1), Mistake Book retrieval and update (Steps 2-2, 2-3, 4-2), test validation against ground truth (Step 3), and composite reward computation with TopVar selection (Step 5-1).}
    \label{fig:detailed_method}
\end{figure*}

Figure~\ref{fig:detailed_method} illustrates the complete training pipeline of \methodname. Given a question from the training dataset, the process proceeds as follows:

\paragraph{Code Generation (Step 1).} The Code LLM generates $M$ candidate solutions $\{C_1, \ldots, C_M\}$ for each question.

\paragraph{Test Generation (Step 2-1).} For each candidate $C_m$, the Test LLM performs white-box generation: it inspects the candidate code and produces $N$ test suites $\{T_{m,1}, \ldots, T_{m,N}\}$, each containing multiple assertion statements.

\paragraph{Mistake Book Retrieval (Step 2-2, 2-3).} The system queries the Mistake Book for historical failures $T^{hist}$ associated with the current question, retrieving tests with the highest failure frequency to form a stable evaluation baseline.

\paragraph{Test Validation (Step 3).} Each generated test suite is executed against the ground-truth code $C_{GT}$ to compute validity scores. Tests with incorrect predictions are corrected using oracle outputs and retained to enrich coverage.

\paragraph{Code Evaluation (Step 4-1).} Candidate solutions are evaluated against the combined test set (newly generated tests plus historical failures). Pass rates determine the adversarial scores for both models.

\paragraph{Mistake Book Update (Step 4-2).} The Mistake Book is dynamically updated: newly failed tests are added with frequency 1, while historical tests that now pass have their frequency decremented or are removed.

\paragraph{Reward Computation (Step 5).} The Code LLM receives rewards based on pass rates. The Test LLM receives composite rewards balancing validity and adversarial difficulty: $R_T = \alpha \cdot \text{Valid} + (1-\alpha) \cdot \text{Adv}$. To balance training compute, only the test suite group with highest reward variance (TopVar) is selected for the Test LLM update.

\label{appendix:method}

\section{Evaluation Metrics Definition}
\label{appendix:metrics}

\subsection{Avg}
\textbf{Avg} represents the macro-average of the code generation performance across all evaluated benchmarks. It is calculated as the arithmetic mean of the avg@32 scores on HumanEval$^+$, MBPP$^+$, and BigCodeBench, providing a single scalar to reflect the model's overall coding capability.

\subsection{Mul}
\textbf{Mul} is designed to evaluate the \textit{yield of high-quality tests} by combining validity and adversarial difficulty. As described in UnLeakedTestBench~\citep{huang2025benchmarking}, the calculation of mutation score ($mut@k$) is inherently conditional on the generated tests first being valid (i.e., compiling and passing the original code).

This dependency creates a trade-off between $pass@k$ and $mut@k$:
\begin{itemize}
    \item A model might achieve a high $pass@k$ by generating trivial, low-complexity tests that lack fault-detection capability (resulting in a low $mut@k$).
    \item Conversely, a model might achieve a high $mut@k$ by generating highly complex tests, even if only a small fraction of them are valid (resulting in a low $pass@k$). This can occur if the metric evaluation considers the ``best" tests from a sparse set of valid candidates.
\end{itemize}

To address this, we define \textbf{Mul} as:
\begin{equation}
    Mul = pass@k \times mut@k
\end{equation}
By multiplying these two metrics, \textbf{Mul} normalizes the mutation score against the total sampling budget. It reflects the true proportion of generated tests that are \textbf{both} valid and adversarial, ensuring that a high score requires the model to simultaneously maintain high generation quality and strong fault-detection capability.

\section{Limitations and Future Work}
\label{appendix:limitation}

\paragraph{Dependence on Ground Truth Code.} \methodname requires ground truth solutions to validate generated tests during training. This limits applicability to domains where reference implementations are unavailable. Future work could explore using consensus among multiple code samples or execution-based validation to relax this requirement.

\paragraph{Test Format Constraints.} The current framework requires tests in assertion format for reliable extraction and execution. This excludes more complex testing scenarios such as stateful interactions, I/O-based testing, or property-based testing. Extending \methodname to support diverse test formats would broaden its applicability.
\paragraph{Generalization to Other Domains.} We evaluate \methodname exclusively on Python function-level code generation. Whether the adversarial co-evolution dynamics transfer to other programming languages, longer code contexts, or non-code reasoning tasks (e.g., mathematical theorem proving with verifiers) remains an open question for future investigation.

\end{document}